\documentclass[journal]{IEEEtran}
\pdfoutput=1
\usepackage{amsmath,amssymb,amsfonts}
\usepackage{hyperref}
\usepackage{cite}
\usepackage{graphicx}
\usepackage{float}
\usepackage{subfig}
\usepackage{makecell}
\usepackage[linesnumbered,ruled,vlined]{algorithm2e}
\usepackage{booktabs}
\usepackage{multirow}
\usepackage{xcolor}
\usepackage{ragged2e}
\usepackage{threeparttable}
\usepackage{caption}
\usepackage[justification=centering]{caption}
\DeclareSubrefFormat{subsimple}{#1(#2)}

\begin{document}
\bstctlcite{IEEEexample:BSTcontrol}
%\title{Evolutionary Generative Adversarial Networks Based on New Fitness Function and Generic Crossover Operator}
\title{IE-GAN: An Improved Evolutionary Generative Adversarial Network Using a New Fitness Function and a Generic Crossover Operator}

\author{Junjie~Li,
		Jingyao~Li,
		Wenbo~Zhou,
        and~Shuai~Lü% <-this % stops a space
\thanks{Manuscript received August 7, 2021; revised October 31, 2022. This work was supported by the Natural Science Research Foundation of Jilin Province of China under Grant No. 20180101053JC; the National Key R\&D Program of China under Grant No. 2017YFB1003103; and the National Natural Science Foundation of China under Grant No. 61300049. \textit{(Corresponding author: Shuai~Lü.)}}% revised August 26, 2015% <-this % stops a space
\thanks{Junjie~Li, Jingyao~Li, and Shuai~Lü are with the Key Laboratory of Symbolic Computation and Knowledge Engineering (Jilin University), Ministry of Education, China; College of Computer Science and Technology, Jilin University, China (e-mail: \href{mailto:junjiel18@mails.jlu.edu.cn}{junjiel18@mails.jlu.edu.cn}; \href{mailto:jingyao18@mails.jlu.edu.cn}{jingyao18@mails.jlu.edu.cn}; \href{mailto:lus@jlu.edu.cn}{lus@jlu.edu.cn}).} %<-this % stops a space
\thanks{Wenbo~Zhou is with the School of Information Science and Technology, Northeast Normal University, China (e-mail: \href{mailto:zhouwb17@mails.jlu.edu.cn}{zhouwb17@mails.jlu.edu.cn}).}}

\markboth{IEEE Transactions on Neural Networks and Learning Systems,~Vol.~XX, No.~X, XXXXXX~2022}%
{Shell \MakeLowercase{\textit{et al.}}: Bare Demo of IEEEtran.cls for IEEE Journals}

\maketitle

\begin{abstract}
The training of generative adversarial networks (GANs) is usually vulnerable to mode collapse and vanishing gradients. The evolutionary generative adversarial network (E-GAN) attempts to {alleviate these issues by optimizing the learning strategy with multiple loss functions. It {uses} a learning-based evolutionary framework, which develops new mutation operators {specifically} for general deep neural networks.} However, {the evaluation mechanism in the fitness function of E-GAN cannot truly reflect the adaptability of individuals to their environment, leading to an inaccurate assessment of the diversity of individuals.} Moreover, {the evolution step of E-GAN only contains mutation operators without considering the crossover operator jointly, isolating the superior characteristics among individuals.} {To address these issues}, we propose a{n improved E-GAN framework called} IE-GAN, which introduces a new fitness function and {a} generic crossover operator. {In particular, the proposed} fitness function{, from an objective perspective,} can {model} the evolutionary {process} of individuals more {accurately}. {The crossover operator, which has been commonly adopted in evolutionary algorithms, can enable offspring to imitate the superior gene expression of their parents through} knowledge distillation. Experiments on various datasets demonstrate the effectiveness of {our proposed} IE-GAN in terms of the quality of {the} generated samples and time efficiency.
\end{abstract}


\IEEEpeerreviewmaketitle

\section{Introduction}
\IEEEPARstart{G}{enerative} adversarial networks (GANs) \cite{DBLP:conf/nips/GoodfellowPMXWOCB14}{, as} deep learning models{, have been} {widely used in synthesis tasks \cite{DBLP:journals/ieeejas/ZhangWM22, DBLP:journals/ieeejas/LiTC21}.} 
{The losses of GANs can be {exploited} by adaptation learning methods to bridge {the gap between} cross-domain distributions \cite{DBLP:journals/tnn/KangYZZA21}.}
However, the training of GANs is {notoriously difficult, as it commonly encounters several} problems{,} such as mode collapse and vanishing gradients \cite{DBLP:conf/iclr/ArjovskyB17}. 
To {address} these {issues}, evolutionary computation has been introduced {to stabilize the training processes of GANs}. {Specifically, evolutionary computation techniques are utilized to perform the selection of generators \cite{DBLP:journals/tec/WangXYT19}, discriminators \cite{DBLP:journals/tec/ChenWXYPCD21}, or even input noise \cite{DBLP:conf/accv/RoziereTHLRZT20} by using} multiple individuals{, which form a population in the selection process}. {A} single individual encoded as {a} genotype {can represent} a neural network in an abstract way. {Actually}, there {exist} many different types of encoded genes, such as network {structures} and weights. {A} genotype is {often} bijectively mapped to the phenotype of {an} individual. The choice of individuals is based on the evaluation of the phenotype{, with the goal of producing} a better solution {in} the next generation {of the population}. In this {research}, we refer to the combination of {GANs} and evolutionary computation as evolutionary {GANs}. 

{The e}volutionary generative adversarial {network} (E-GAN) \cite{DBLP:journals/tec/WangXYT19} is a {typical} evolutionary {GAN, that} maintains a generator population {to generate} individuals with different loss functions{, which are referred to} as mutation operators. In each {evolution} step {in E-GAN}, individuals are evaluated in terms of {their} quality and diversity. {Due to the success of E-GAN, many researchers have tried to further improve the performance of} E-GAN. For example, {the} cooperative dual evolution{-based} GAN (CDE-GAN) {\cite{DBLP:journals/tec/ChenWXYPCD21}} expands the concept of {a} population to {the} discriminator and uses a soft mechanism to connect two populations {that compete with each other during the evolution process}. {Moreover,} Mu \textit{et al.} {\cite{mu2020enhanced}} {defined} a novel mutation operator, i.e., a distribution indicating realness, {to replace the original mutation operator in E-GAN}. {At the same time, the multi-objective evolutionary GAN (MO-EGAN) {\cite{DBLP:conf/gecco/BaiolettiCBP20}} and smart MO-EGAN (SMO-EGAN) {\cite{DBLP:conf/cec/BaiolettiBPC21}} consider quality and diversity as conflicting objectives, and {they} model the evaluation of generators as a multi-objective problem.} {In addition, mutation spatial GANs} (Mustangs) {\cite{DBLP:conf/gecco/ToutouhHO19}} applies different loss functions to {a} competitive co-evolutionary algorithm. {In addition to} the E-GAN-based methods, there are many other studies on evolutionary {GANs} \cite{DBLP:conf/aaai/LiuWL20,DBLP:conf/gecco/Costa0CM19,DBLP:conf/gecco/Costa0CM20,DBLP:conf/accv/RoziereTHLRZT20,DBLP:conf/gecco/GarciarenaSM18}. Nevertheless, each of these methods {usually has weaknesses with respect to the individual evaluation process or the crossover operator}.

{An evolutionary strategy requires an appropriate fitness function to assess the adaptability of individuals to their environment. The fitness function models the researcher's expectation of the evolutionary trend of a population and is an essential guarantee of the effectiveness of an evolutionary strategy. Specifically, for the generative model, {a} fitness function evaluates the samples generated by individuals. Nevertheless}, {determining} how to measure the diversity of the generated samples is also a controversial problem. {The} Fr\'echet inception distance (FID) \cite{DBLP:conf/nips/HeuselRUNH17}{, which} can reflect samples' diversity{, is often utilized as a typical metric for generative performance during testing. The resource requirements of {the} FID{, however,} make it unsuitable {for use} in the training phase of methods with a large number of iterations.} {As a result,} E-GAN {uses} a discriminator-based diversity metric{. However, according to our observations from experiments, we found that this metric cannot be as effective as expected for E-GAN.}

{On the other hand,} there are two common but crucial variation operators in evolutionary computation, i.e., mutation and crossover. The mutation operator produces new {characteristics}, {while} the crossover operator is used to combine the best {characteristics} of multiple individuals. {Currently}, evolutionary {GANs} rarely {consider} the crossover operator. Although Garciarena \textit{et al.} applied the crossover operator in \cite{DBLP:conf/gecco/GarciarenaSM18}, {their defined} crossover operator is {rather} difficult to {adapt} to other methods. 
{In \cite{DBLP:conf/gecco/GarciarenaSM18}, pairs of generators and discriminators {regarded} as evolv{ing} individuals are crossed by exchanging paired objects. The cross-variance implementation dictates that the operator is only applicable to methods that have pairwise generators and discriminators.}
{Moreover, this crossover strategy can only expand the population diversity; it cannot make the offspring collect the superior {characteristics} of their parents, {due to the fact that} the neural network does not reflect gene dominance or recessiveness, which play an important role in biological inheritance.}

{In this research, we propose an improved E-GAN framework called IE-GAN to address these existing shortcomings of evolutionary GANs. Specifically, IE-GAN introduces an objective fitness function and a generic crossover operator to evolutionary GANs. Our proposed fitness function considers the perspective of individual phenotypes,  which is more accurate while balancing the time cost; the diversity part of the fitness function is based on gene expression without relying on other training entities, and the quality part of the fitness function is adjusted accordingly. The $E$-filtered knowledge distillation crossover in IE-GAN also starts from the phenotype rather than directly manipulating genes like $n$-point crossover, a typical crossover operator for evolutionary computation. This learning-based crossover operator allows offspring to imitate the gene expression of their parents through knowledge distillation and preserve the learned knowledge in their genotypes, which protects the training stability of the GAN. In essence, IE-GAN directs the exploration direction of possible solutions in the parameter space more accurately with fewer resources while exploiting the valuable parts of multiple trials. Extensive experiments on multiple datasets show that our method has advantages over state-of-the-art methods in terms of both {the} generative performance and training time, and {it} can alleviate mode collapse {and} vanishing gradients; it is also quite robust.}

{To summarize, the main} contributions of this paper are as follows:
\begin{itemize}
	\item We analyze and experimentally verify the imperfection of E-GAN. {The fitness function of E-GAN has difficulty modeling the environmental pressure on the population, and therefore, it cannot guide the population in the desired direction. E-GAN also lacks a crossover operator and forgets the valuable experience of individuals who did not survive.}
	\item {We propose a concise but efficient fitness function that can accurately reflect the individuals' adaptation to the environment. Unlike previous functions, it can objectively measure the variance of the generated samples while requiring less time.}
	\item {We introduce a generic learning-based crossover operator that clusters the superior characteristics of multiple networks irrespective of specific network implementations. To the best of our knowledge, this is the first crossover operator that is widely applicable to methods that consider deep networks as individuals.}
	\item We design and implement a framework called IE-GAN with our crossover operator and fitness function. {Experiments show that our method has significant advantages in terms of the training stability, generative performance, and training time. This framework is also robust and can be easily applied to other evolutionary GANs.}
\end{itemize}

{The remainder of this paper is organized as follows. Section \ref{Related Work} reviews the main relevant studies. Section \ref{Preliminaries} presents a priori knowledge. Section \ref{Method} elaborates on the proposed IE-GAN. Section \ref{Experiments} describes the experiments. Section \ref{Conclusion} summarizes our work.}

\section{Related Work}
\label{Related Work}

\subsection{E-GAN}
E-GAN \cite{DBLP:journals/tec/WangXYT19} {uses} an adversarial framework {with} a discriminator and a population of generators{ to improve the stability and generative capability of GANs}. Specifically, it assumes that the generators no longer exist as individuals{; instead, they are} a population {that competes} with {the} discriminator. From an evolutionary point of view, the discriminator can be considered a changing environment during the evolutionary process. 

In each {evolution} step of E-GAN, the evolution of the generator consists of three steps: variation, evaluation, and selection.
The variation includes only {the} mutation operation, and different loss functions are chosen as mutation operators to produce different offspring. Then, the generative performance of the generated offspring needs to be evaluated and quantified {using} the corresponding fitness {value}. We will {call} the fitness function in \cite{DBLP:journals/tec/WangXYT19} $F^{E-GAN}$ in this paper to avoid confusion. After measuring the generative performance of all offspring, according to the principle of survival of the fittest, the updated generators are selected as the parents for a new round of training.

{The mutation operator used by E-GAN is more suitable for deep neural networks than random mutations that imitate organisms. A framework based on gradient learning rather than randomness allows the population to remain small and allows the system to converge to meaningful results relatively quickly. This evolutionary strategy guided by loss functions is more conducive to system stability. Meanwhile, this evolutionary process derived from learning is easily extended to other algorithms and tasks.}
{In view of the above advantages of E-GAN, Li \textit{et al.}} {created an} E-GAN with crossover (CE-GAN) in \cite{DBLP:conf/ijcnn/LiZGL21}, which is {a} preliminary step {for} this work.

\subsection{Knowledge Distillation}
Knowledge distillation is a model compression method {and} training method based on the {teacher-student network} in which the knowledge contained in a trained model is distilled and applied to the target model {\cite{DBLP:journals/corr/HintonVD15}}. The target model is often a smaller model that is compressed and has better generalizability. The target model can learn to match any layer of the trained model, {including intermediate feature maps of different sizes \cite{DBLP:journals/tip/HuangYZLGC22}.} 
Hard targets learn{ from} the output, which reduces {the} training time but increases the possibility of overfitting. {S}oft targets learn from logits, which contain more descriptive information about the samples and enable the target network to have better generalization {capabilities} \cite{DBLP:journals/corr/HintonVD15}.

As a means of compressing networks, knowledge distillation has been applied to GAN{s} \cite{DBLP:conf/cvpr/LiLDLZH20,DBLP:journals/corr/abs-1902-00159}. 
Proximal distilled evolutionary reinforcement learning (PDERL) \cite{DBLP:conf/aaai/BodnarDL20} utilizes this technique to optimize evolutionary reinforcement learning (ERL) \cite{DBLP:conf/nips/KhadkaT18}. It {utilized} a {learning-based} crossover operator {called the} $Q$-filtered behavior distillation crossover {operator}. {E}xperiments demonstrate that in reinforcement learning (RL), {this} operator can ensure that offspring inherit the {behaviors} of their parents. Thus, $Q$-filtered {behavior} distillation crossover will not cause destructive interference like $n$-point crossover. Moreover, $Q$-filtered {behavior} distillation crossover as a learning-based variation can be easily accelerated using graphics processing units (GPUs){,} and {it} is useful even {for} deeper networks. {The crossover operator proposed in this paper is also implemented based on knowledge distillation.}

\subsection{Gradient Penalty}
{The g}radient penalty (GP) \cite{DBLP:conf/nips/GulrajaniAADC17} achieves the Lipschitz constraint required {by the W}asserstein GAN (WGAN) \cite{DBLP:conf/icml/ArjovskyCB17}. It replaces the weight clipping of WGAN and effectively avoids the concentration of weights on the clipping threshold, so the network will not push weights towards two values. {Additionally}, because it eliminates the setting of the clipping threshold, the network will not suffer from vanishing or exploding gradient{s} because of {an} unreasonable threshold setting.
{The application space of {the} GP {has been} extended from the image space to separable Banach spaces \cite{DBLP:conf/nips/AdlerL18}.}

Although {the} GP was originally used to improve WGAN, {it can be part of other frameworks as a regularization term \cite{DBLP:journals/spl/YangZXGQ21}. {The} GP is orthogonal to E-GAN and can regularize the discriminator to support {the} updating {of} generators \cite{DBLP:journals/tec/WangXYT19}.}
{Experiments show that our IE-GAN can also benefit from {the} GP.}

\section{Preliminaries}
\label{Preliminaries}

\subsection{Notation}
We refer to the symbols used in \cite{DBLP:conf/iclr/Jolicoeur-Martineau19,DBLP:journals/tec/WangXYT19}. GANs can { generally} be formulated as {follows}:
\begin{equation}
	{\mathcal{L}_{D}}=\mathbb{E}_{x \sim p_{data}}[f_{1}(C(x))]+\mathbb{E}_{z \sim p_{z}}[f_{2}(C(h(T(z))))],
\end{equation}
\begin{equation}
	{\mathcal{L}_{G}}=\mathbb{E}_{x \sim p_{data}}[g_{1}(C(x))]+\mathbb{E}_{z \sim p_{z}}[g_{2}(C(h(T(z))))],
\end{equation}
where $f_1$, $f_2$, $g_1$, {and} $g_2$ are scalar-to-scalar functions, and the GAN and its variants with different loss functions {use different versions of} these functions. $p_{data}$ is the distribution of real data, {and} $p_z$ is the distribution of {the} sampling noise (usually a uniform or normal distribution). {$T(\cdot)$ is the generative network without {the} final activation layer,} {and $h(\cdot)$ is an activation layer, which} is $tanh(\cdot)$ in {the} deep convolutional GAN (DCGAN) \cite{DBLP:journals/corr/RadfordMC15}. $G(z)=h(T(z))$, where $G(z)$ is the output of the generator and its distribution $p_g$ is expected to be fitted {to} $p_{data}$. {$C(\cdot)$ is the discriminative network without {the} final activation layer.} In most GANs, $C(x)$ can be interpreted as {indicating} how realistic the input data are \cite{DBLP:conf/iclr/Jolicoeur-Martineau19}. In the original GAN \cite{DBLP:conf/nips/GoodfellowPMXWOCB14}, $D(x)=Sigmoid(C(x))$, where $D(x)$ is the output of the discriminative network $D$. 

\subsection{Fitness Function}
An evaluation criterion to measure the quality of individuals is needed in the evolutionary algorithm. Existing methods utilize various fitness functions, such as {the} inverted generational distance (IGD) \cite{DBLP:conf/gecco/GarciarenaSM18}, Koncept512 \cite{DBLP:conf/accv/RoziereTHLRZT20}, $F^{E-GAN}$ and its variants \cite{DBLP:journals/tec/WangXYT19,DBLP:journals/tec/ChenWXYPCD21,mu2020enhanced,DBLP:conf/gecco/BaiolettiCBP20}, and a mixture of {GAN objective functions and the FID \cite{DBLP:conf/gecco/ToutouhHO19,DBLP:conf/gecco/Costa0CM19}}.

{The f}itness function $F^{E-GAN}$ \cite{DBLP:journals/tec/WangXYT19} is {written} as follows:
\begin{equation}
	F^{E-GAN}=F^{E-GAN}_q+\gamma F^{E-GAN}_d,
	\label{eq:E-GAN_F}
\end{equation}
where $\gamma>0$ balances two measurements. 
The output of the discriminator $D$ w.r.t. the generated samples $G(z)$ {is} used as the quality fitness $F^{E-GAN}_q$:
\begin{equation}
	F^{E-GAN}_q=\mathbb{E}_{z \sim p_{z}}[D(G(z))].
	\label{eq:E-GAN_Fq}
\end{equation}
The diversity fitness function $F^{E-GAN}_d$ is formally expressed as
\begin{equation}
	\begin{split}
		F^{E-GAN}_d=-\log \|\nabla_{D}&-\mathbb{E}_{x \sim p_{data}}[\log D(x)]\\
		&-\mathbb{E}_{z \sim p_{z}}[\log (1-D(G(z)))]\|_2.
	\end{split}
	\label{eq:E-GAN_Fd}
\end{equation}
{T}he diversity fitness is the {negative} log-gradient-norm of the loss function of $D$. The logarithm is used to shrink the fluctuation of the gradient norm, which differs from the output of the discriminator $D$ by an order of magnitude, and the amplitude is so large that a simple balance coefficient cannot balance the two {values}.

E-GAN {assumes} that the more discrete the generated samples are, the less likely the discriminator is to change the parameters substantially. $F^{E-GAN}_d$ embodies this principle. {However,} our experiments show that $F^{E-GAN}_d$ not only cannot achieve {this} design purpose but also {causes} obvious {negative} side effects.

\section{Method}
\label{Method}
{This section describes the details of our proposed IE-GAN, especially the {main contributions of the} evaluation {of individuals} and crossover.}

\subsection{IE-GAN Framework}
\begin{figure*}[!t]
	\centering
	\includegraphics[width=0.85\textwidth]{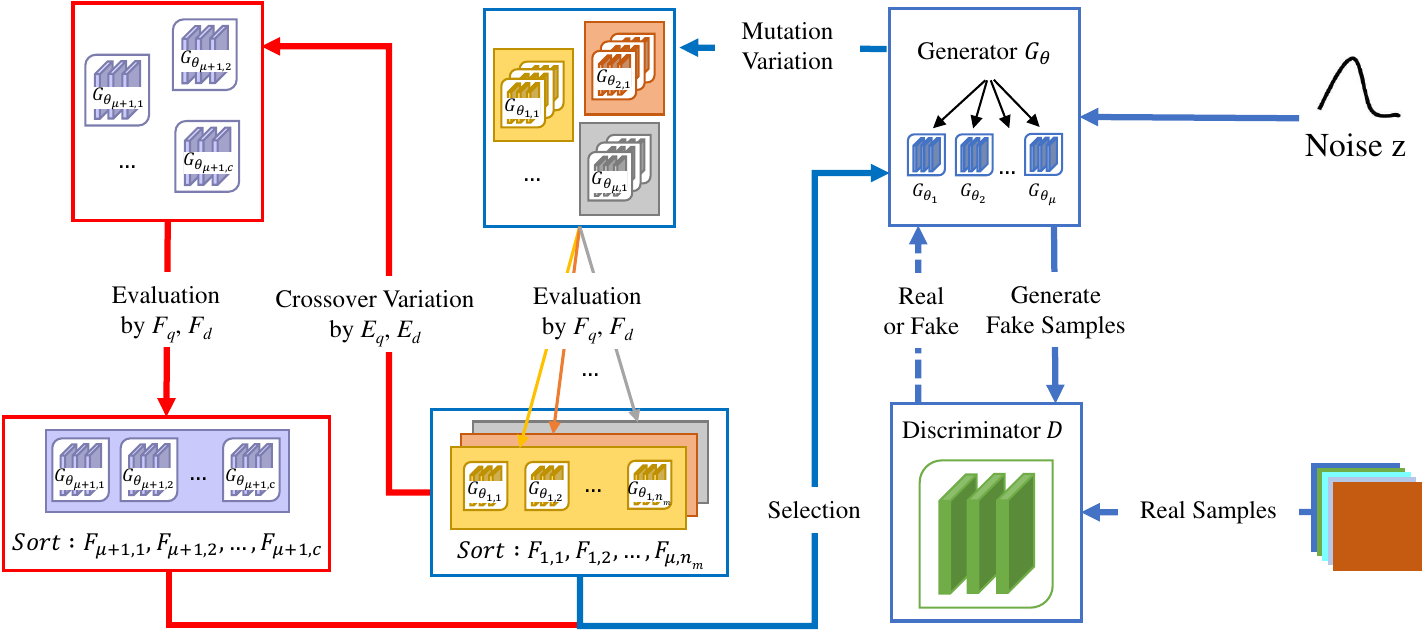}
	\caption{{~The proposed IE-GAN framework. It adds crossover variation to take advantage of the experience of mutation individuals and enables a more efficient and accurate fitness evaluation function. The flow added in the {evolution} step is {shown} in red.}}
	\label{fig:IEGAN}
	\vspace{-0.5cm}
\end{figure*}

The IE-GAN {framework} designed in this paper evolves a population of generators {$\{G_i\}_{i=1}^{\mu}$} in a given dynamic environment (discriminator $D$) and a static environment (diversity fitness function), {where $\mu$ is the size of the population}. Each individual in the population represents a possible solution in the parameter space \cite{DBLP:journals/tec/WangXYT19}. {The} IE-GAN framework is shown in Fig.~\ref{fig:IEGAN}. 

In each {evolution} step, individuals $\{G_{\theta_1}, \cdots, G_{\theta_\mu}\}$ in the population $G_\theta$ act as parents to generate offspring $\{G_{\theta_{1,1}}, G_{\theta_{1,2}}, \cdots, G_{\theta_{\mu,n_m}}\}$ {using} mutation operators, {where $n_m$ is the number of mutation operators}. After evaluating {mutation individuals} using {the} fitness function $F(\cdot)$, {their} environmental fitness is obtained. Then crossover parents are selected according to these fitness scores, and crossover individuals $\{G_{\theta_{\mu+1,1}}, G_{\theta_{\mu+1,2}}, \cdots, G_{\theta_{\mu+1,c}}\}$ are {obtained using the} crossover operator, {where $c$ is the number of crossover individuals}. Next, the fitness of crossover individuals {is evaluated}. Finally, the offspring with high fitness scores are selected as {the} parents of the next generation.

After {the evolution} step, the discriminator $D$ is updated. Thus the dynamic environment also changes with the evolution of {the} population, and {it} is able to provide continuous evolutionary pressure in conjunction with {the} static environment. The objective function of $D$  is the same as that of the original GAN \cite{DBLP:conf/nips/GoodfellowPMXWOCB14}, {except for the additional GP term \cite{DBLP:conf/nips/GulrajaniAADC17},} {i.e.,}
\begin{equation}
	\begin{split}
		{\mathcal{L}^{IE-GAN}_D}=&-\mathbb{E}_{x \sim p_{data}}[\log D(x)]\\
		&-\mathbb{E}_{z \sim p_{z}}[\log (1-D(G(z)))]\\
		&+{\lambda\mathbb{E}_{\hat{x} \sim p_{\hat{x}}}[(\|\nabla_{\hat{x}}{D(\hat{x})}\|_2-1)^2]},
	\end{split}
	\label{eq:Dloss}
\end{equation}
{where the third subterm is a GP term. $p_{\hat{x}}$ is the intermediate distribution between {the} real data and generated data. $\lambda$ is the balance coefficient of {the} GP, which is {set to} $\lambda=10$ as {in} WGAN-GP \cite{DBLP:conf/nips/GulrajaniAADC17}.}

{The} IE-GAN framework implemented in this paper adopts the same three mutation operators as E-GAN: {the} minimax ({in} GAN \cite{DBLP:conf/nips/GoodfellowPMXWOCB14}), heuristic ({in} NS-GAN \cite{DBLP:conf/iclr/ArjovskyB17}), and least-squares ({in} LSGAN \cite{DBLP:conf/iccv/MaoLXLWS17}) {operators}.
The complete algorithm is similar to the framework algorithm in \cite{DBLP:conf/ijcnn/LiZGL21}. 
{In comparison} to the two-player game of traditional GANs, E-GAN {makes it possible} to select different dominant adversarial targets in different training phase{s}. Furthermore, CE-GAN \cite{DBLP:conf/ijcnn/LiZGL21} can integrate the advantages of multiple loss functions in each training {phase} to generate advantageous individuals. Moreover, IE-GAN improves the crossover operator of CE-GAN and is able to guide evolution with a more accurate evaluation function, thereby obtaining a more competitive solution.
It is worth mentioning that the discriminator $D$ is deeply involved in the {entire} framework. It provides gradient information for mutation operators{,} and it {provides} an important measurement {that is used to evaluate} samples and offspring. {Therefore,} the framework can {strongly} benefit from the improvement of $D$.

\subsection{Mutation}
{The evaluation function and the crossover operator of IE-GAN, both of which act on the generated samples, are independent of the network implementation.} {Therefore, the IE-GAN framework is independent of the genes encoded.}
It has no hard and fast rules for the mutation operators, as long as the mutation operators can proliferate the parents into several different mutation individuals. 
{As a result}, the IE-GAN framework{, which does not introduce additional constraints,} can be easily applied to existing evolutionary {GANs} with generators as individuals.

\subsection{Evaluation}
{The} FID is a common generation task metric in {the} literature \cite{DBLP:conf/iclr/BrockDS19, DBLP:conf/cvpr/KarrasLA19, DBLP:conf/cvpr/KarrasLAHLA20, DBLP:journals/ijcv/LeeTMHLSY20, zhou2022coutfitgan, zhou2022learning}{; it reflects both the} quality and diversity of the samples. It can also be used as {a} fitness function to guide evolution in an evolutionary strategy. However, {its} higher time complexity makes it suitable only for methods with {fewer} evolutionary generations{;} otherwise the excessive time cost will make the method useless.

$F^{E-GAN}$ calculates the environmental fitness score of an individual by evaluating the $l$ samples it generates. It splits the assessment metric into two aspects: {the} quality fitness score and {the} diversity fitness score. However, equation \eqref{eq:E-GAN_Fd} cannot fully express the diversity fitness score. 
In fact, $F^{E-GAN}_d$ is affected {not only} by diversity but also by quality. As an example, if the generated sample distribution hardly overlaps with the true sample distribution, $D$ can {almost accurately distinguish the two types of samples}. When the $D$ network hardly changes the parameters, the evaluation object will become unduly competitive. To put it simply, low{-}quality samples will {receive} high diversity fitness scores. This is the problem: $F^{E-GAN}_d$ not only fails to reflect diversity but also hinders the quality evaluation of $F^{E-GAN}_q$, which makes {the} training vulnerable to {the} vanishing gradients {problem} when the balance coefficient is not appropriate.
Furthermore, the nonlinear scaling of $F^{E-GAN}_d$ amplifies its negative effects, and even the most careful adjustment of the balance coefficients can easily {cause} the evaluation results {to be} dominated by a single submetric.
In addition, $F^{E-GAN}_d$ requires additional computation for the gradient information of {the} discriminator, which increases the time cost significantly. 

\begin{figure}[!t]
	\centering
	\includegraphics[width=0.45\textwidth]{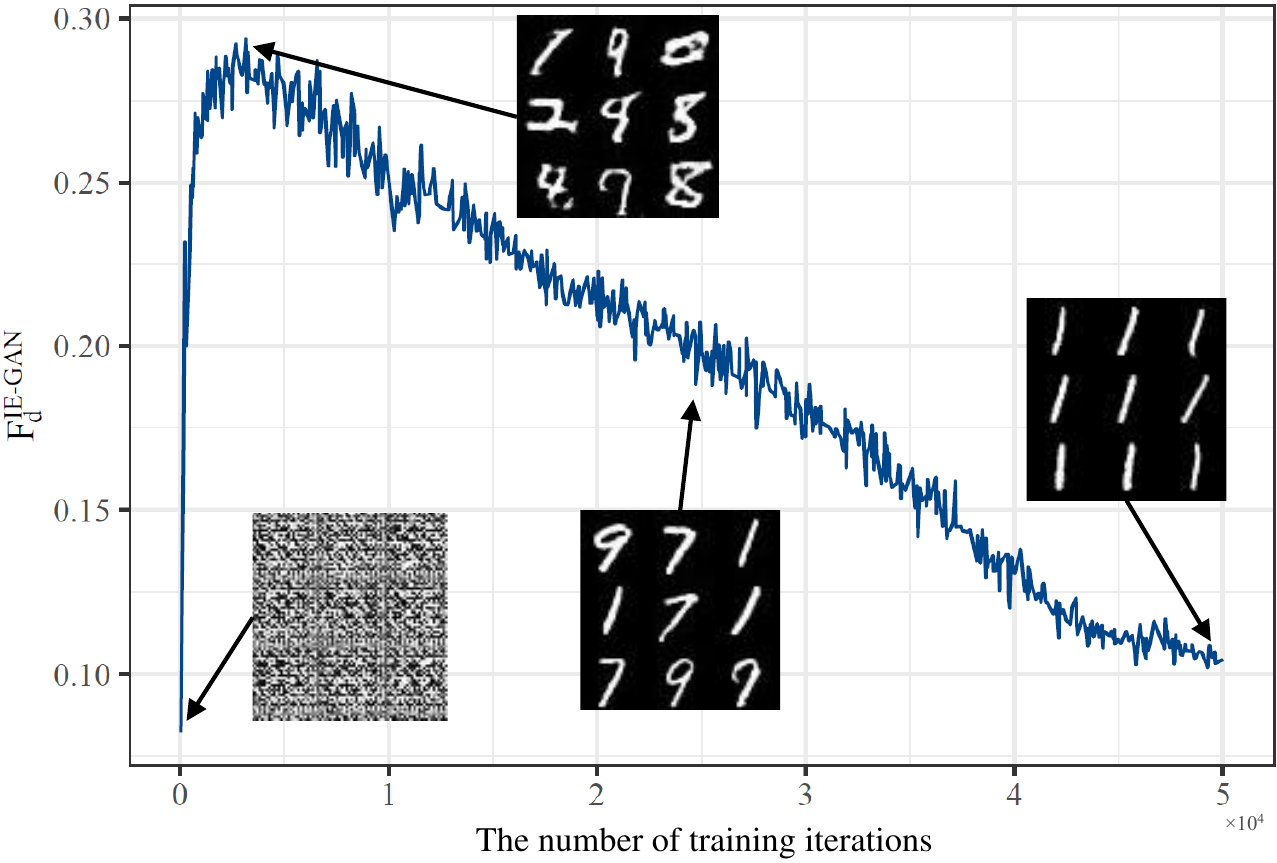}
	\caption{{~$F^{IE-GAN}_d$ curves and samples at different training stages {for} MNIST.}}
	\label{fig:Fd&Samples}
	\vspace{-0.5cm}
\end{figure}

{Rather than relying on synchronous training discriminators like $F^{E-GAN}_d$, we expect the diversity fitness function to measure individuals objectively. Instead of indirectly evaluating the diversity of the generators using a discriminator, we directly evaluate the distributional distance of the samples produced by the generators.}
{Therefore}, we propose a new diversity fitness function, which estimates {the} diversity in terms of {the} mean absolute error (MAE) between samples. The generative diversity of individuals is measured by {the} sample dissimilarity. Formally, our diversity fitness function is defined as follows:
\begin{equation}
	F^{IE-GAN}_d=\dfrac{1}{n_e}\sum_{i=1}^{n_e}\mathbb{E}_{z_1,z_i \sim p_{z}}[ \|G(z_1)-G(z_i)\|_1],
	\label{eq:IE-GAN_Fd}
\end{equation}
where $n_e$ refers to the number of times each sample is compared with other samples. It is the mean w.r.t. the MAE of each sample. 
{The fitness function is easy to compute, which is {helpful} to algorithms that require a large number of iterations. The function is also robust and insensitive to outliers, which {keeps} the global perspective {from being} overly influenced by local differences. Moreover,}
the MAE of a single sample can also be considered as the evaluation of the sample itself, which helps the subsequent crossover operator to make full use of the evaluated samples. As shown in Fig.~\ref{fig:Fd&Samples}, experiments on the MNIST\footnote{http://yann.lecun.com/exdb/mnist/} dataset demonstrate that our fitness function can objectively reflect the sample diversity.

{Additionally, our quality fitness function also differs from equation \eqref{eq:E-GAN_Fq} by replacing $D(\cdot)$ in $F^{E-GAN}_q$ with $C(\cdot)$. The Sigmoid layer does not change the sorting but affects the output values. When combined with the diversity component, the presence or absence of the Sigmoid layer may lead to changes in the fitness sorting. Given that $C(x)$ can represent the truthfulness of the data $x$ \cite{DBLP:conf/iclr/Jolicoeur-Martineau19}, we believe that $C(\cdot)$ is {a} more suitable function {to use} to assess the quality fitness.}
The formal definition of our quality fitness function is as follows:
\begin{equation}
	F^{IE-GAN}_q=\mathbb{E}_{z \sim p_z}[C(G(z))].
	\label{eq:IE-GAN_Fq}
\end{equation}
Its essence is the evaluation of the sample quality {using} the discriminator. It can measure a single sample without additional operations, {and} therefore it can be used to filter samples in {the} crossover {operation}.

{By combining} the above two fitness functions, the fitness function of the IE-GAN framework can be obtained:
\begin{equation}
	F^{IE-GAN}=F^{IE-GAN}_q+\gamma_1 F^{IE-GAN}_d,
	\label{eq:IE-GAN_F}
\end{equation}
where $\gamma_1>0$ balances the two metrics. A higher $F^{IE-GAN}$ {value} indicates that the evaluated object has {a} better generative performance.

\subsection{Crossover}
\begin{figure}[!t]
	\centering
	\includegraphics[width=0.45\textwidth]{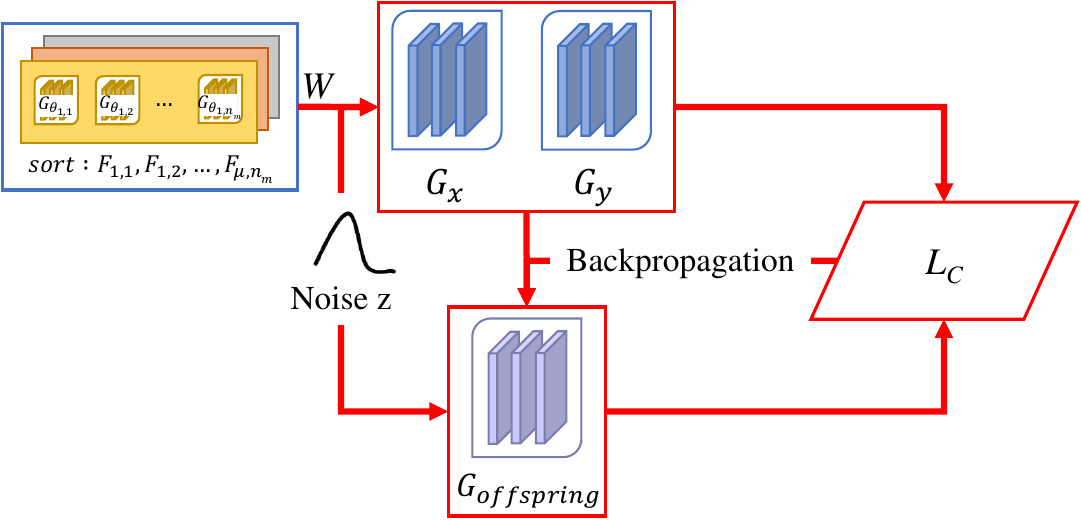}
	\caption{{~The birth of a crossover offspring. Crossover parent pairs are selected from the mutation offspring according to the function $W$. The basic network of the crossover offspring is derived from the parents. For three networks with the same noise input, the offspring imitates learning {through the} backpropagation of the loss function $\mathcal{L_C}$.}}
	\label{fig:Crossover}
	\vspace{-0.5cm}
\end{figure}

In this subsection, we present the $E$-filtered knowledge distillation crossover {operator} w.r.t. the $Q$-filtered behavior distillation crossover {operator} \cite{DBLP:conf/aaai/BodnarDL20}. Due to the difference between GAN{s} and RL, our crossover operator differs from {this previous crossover operator} in some aspects{,} including sample filtering and {basic} network. 
Fig.~\ref{fig:Crossover} illustrates the flow of a crossover {operation}.

The crossover operation of a pair of parent networks is as follows:
\begin{itemize}
	\item The model with a higher fitness score among the parents is used as the basis of the offspring model, and the results generated by the parents with the same inputs are used as experiences for the offspring to {learn from}. 
	\item {T}he knowledge of {the} parents {is transferred} to {the} offspring {through} knowledge distillation. 
\end{itemize}
Note that the parents mentioned in {the definition of} the crossover {operation are} the mutation individuals generated by the mutation operators rather than the original parents of each generation.

The outputs of parents for the same inputs are likely to be completely different. {Thus,} a natural question is {the following:} how {can} the offspring imitate both networks at the same time? The solution is that the {offspring} choose the better output to imitate for each input. The definition of {whether a} sample $x$ {is good or bad} is left to $E(x)$. The higher the $E(x)$ {value}, the more worthy of study {a given output is}. $E(x)$ is defined as follows:
\begin{equation}
	E=E_{q}+\gamma_2 E_{d},
	\label{eq:E}
\end{equation}
{where $\gamma_2>0$ is a parameter used to balance the two metrics.} This evaluation also considers both {the} quality and diversity of samples{:}
\begin{equation}
	E_{q}=C(x),
	\label{eq:Eq}
\end{equation}
\begin{equation}
	E_{d}=\dfrac{1}{n_e}\sum_{i=1}^{n_e} {\mathbb{E}_{x \sim p_g}}[\|x-x_i\|_1],
	\label{eq:Ed}
\end{equation}
where $x_i$ is a sample from the same batch as $x$.

The formal representation of the $E$-filtered knowledge distillation loss used to train the offspring is as follows:
\begin{equation}
	\begin{split}
		{\mathcal{L}_{C}}&=\mathbb{E}_{E(G_{x}(z_{i}))>E(G_{y}(z_{i}))|z_i \sim p_{z}}[\|T_{offspring}(z_{i})-T_{x}(z_{i})\|^{2}] \\
		&+\mathbb{E}_{E(G_{y}(z_{j}))>E(G_{x}(z_{j}))|z_j \sim p_{z}}[\|T_{offspring}(z_{j})-T_{y}(z_{j})\|^{2}],
	\end{split}
\end{equation}
where $G_x$ and $G_y$ represent the generative strategies of {the} parents. $T_x(\cdot)$ and $T_y(\cdot)$ are the inactive outputs of their corresponding generators, and $T_{offspring}(\cdot)$ represents an inactive output of {the offspring}. We believe that fitting with soft targets help{s} the offspring inherit more information.
The offspring network learns the knowledge of {the} parents by minimizing the function {$\mathcal{L_C}$} during training. The whole procedure of the crossover operator is similar to {the} $C$-filtered knowledge distillation crossover algorithm in \cite{DBLP:conf/ijcnn/LiZGL21}. It {requires} the hyperparameter crossover batch size{, which is denoted by} $n$.

There is still a problem: how {should the algorithm} select parents from {among} the multiple networks for crossover? We use the greedy strategy score function $W$ defined in \cite{DBLP:conf/aaai/BodnarDL20}:
\begin{equation}
	W=F(G_x)+F(G_y)
\end{equation}
This strategy usually results in the selection of better individuals, which ensures the excellence of the resulting offspring and increases the stability of the population \cite{DBLP:conf/aaai/BodnarDL20}.
Since crossover parents and the initial network of crossover offspring need to be selected based on individual evaluation, it is necessary to evaluate mutation individuals before crossover. Given that the assessment{s} of {both} individuals and samples are considered in terms of {the} quality and diversity, the samples generated and the fitness scores obtained in the evaluation substage can be reused in the crossover substage.

This operator can act on the phenotype space \cite{DBLP:journals/corr/abs-1902-00159} or the extracted intermediate features \cite{DBLP:conf/cvpr/LiLDLZH20}. Knowledge distillation is {independent} of the topological differences between {the} teacher network and student network, but if learning from feature extraction {is taking place}, it is a challenge to select matches from the constantly changing network architecture for an evolutionary strategy that encodes {the} topology as genes. In view of this, we choose to perform imitation learning from the end of network so that {this method} can be applied to a more general framework. 

The crossover operator proposed in this paper is somewhat different from crossover variation {in the biological sense}. In biology, the phenotypic {characteristics} of offspring are determined by gene expression, and the dominance of inherited {characteristics} is reflected by the dominance of genes (usually the {characteristics} represented by dominant genes are better). In short, gene crossover determines {the} inheritance {of characteristics}. The basis for {realising} our crossover operator is learning the expression of {superior} parental {characteristics} in the phenotype space and inscribing the knowledge in the network weights. The superior {characteristics} are judged by the offspring themselves. The entire process does not require the participation of genes. {Because} of this gene-independence, the $E$-filtered knowledge distillation crossover operator does not care about the genes encoded by the individuals of the population.

\subsection{Selection}
There are two {issues related to} the selection of the next generation {of} parents: whether {or not} the mutation individuals are still candidates{, a}nd whether {or not} the current parents are candidates{.}
For the former {issue}, we have experimentally demonstrated that only using crossover offspring will not bring significant gains, and doing so does not reduce the computational complexity.
Therefore, mutation offspring should not be excluded from selection.
As for the latter issue, existing studies use updated discriminators  to re-evaluate current parents and treat them as parent candidates for the next generation \cite{DBLP:journals/tec/ChenWXYPCD21,DBLP:conf/gecco/BaiolettiCBP20}. Experiments on our framework show that this strategy enhances {the} training stability. {However,} it will slow down convergence, {negatively affect the final result,} and increase {the} computational effort. {Therefore,} we do not consider transferring the current parents to the next generation.

\section{Experiments}
\label{Experiments}
To verify the effectiveness of the IE-GAN framework, we {conducted extensive experiments} with its algorithmic implementation on several generative tasks{,} and {we} present the results in this section. The experiments show that IE-GAN can achieve {a} competitive performance with high efficiency. In addition, we analyze some details of the method.

\subsection{Experimental Configuration}
We conduct experiments on two synthetic datasets and two real-world image datasets{,} CIFAR-10 \cite{krizhevsky2009learning} and CelebA \cite{DBLP:conf/iccv/LiuLWT15}. The two synthetic datasets are a 2D mixture of {eight} Gaussians arranged in a circle and a 2D mixture of 25 Gaussians arranged in a grid. Because the distribution{s} of {the} generated data and target data can be visualized, such toy datasets are often used to demonstrate the mode collapse of {a} method. CIFAR-10 is a dataset for identifying universal objects {that contains ten} categories of RGB color images {of} objects in the real world. The images in {this dataset} are noisy, {and} the proportions and features of the objects are also different, which {makes} recognition {difficult}. The size of the images is $32\!\times\!32$. CelebA is a large-scale face attribute dataset with a large number of celebrity images. The images in this dataset {have} pose variations and background clutter.

For comparison purposes, the generator and discriminator {networks for} CIFAR-10 are the same as {those in} E-GAN \cite{DBLP:journals/tec/WangXYT19}{; they} are fine-tuned DCGAN networks \cite{DBLP:journals/corr/RadfordMC15}. As for CelebA, we adopt the official PyTorch implementation\footnote{https://github.com/pytorch/examples/tree/master/dcgan} of the DCGAN architecture. To cope with the complex dataset CelebA, we apply more channels in the convolutional layers and transposed convolutional layers. In addition, we choose {the} multilayer perceptron (MLP) as the model architecture for the toy experiments. The target distribution of {the} synthetic data and the corresponding network architectures in the experiments are taken from the third-party PyTorch implementation\footnote{https://github.com/caogang/wgan-gp} of WGAN-GP. The model architecture applied in this paper is shown in detail in Table~\ref{tab:Architectures}. {In addition to the network architectures, the GP settings of the two types of datasets are also different. In order to better reflect the performance differences of each method on synthetic datasets, experiments on {the eight} Gaussians and 25 Gaussians do not use {the} GP term.}

\begin{table*}[!t]
	\caption{Architectures of the Generative and Discriminative Networks}
	\label{tab:Architectures}
	\centering
	\begin{tabular}{l|l}			
		\hline
		\textbf{Generative Network}	&	\textbf{Discriminative Network} \\
		
		\hline
		\textbf{E-GAN}	&	\\
		\textbf{Input}: Noise $z \sim p_z$, 100										&	
		\textbf{Input}: Image ($32 \times 32 \times 3$)							\\
		$[$layer 1$]$  Transposed Convolution (4, 4, 512), stride=1; \textit{ReLU};	&
		$[$layer 1$]$  Convolution (4, 4, 128), stride=2; \textit{LeakyReLU};		\\
		$[$layer 2$]$  Transposed Convolution (4, 4, 256), stride=2; \textit{ReLU};	&
		$[$layer 2$]$  Convolution (4, 4, 256), stride=2; Batchnorm; \textit{LeakyReLU};\\
		$[$layer 3$]$  Transposed Convolution (4, 4, 128), stride=2; \textit{ReLU};	&
		$[$layer 3$]$  Convolution (4, 4, 512), stride=2; Batchnorm; \textit{LeakyReLU};\\
		$[$layer 4$]$  Transposed Convolution (4, 4, 3), stride=2; \textit{Tanh};	&
		$[$layer 4$]$  Fully Connected (1); \textit{Sigmoid};						\\
		\textbf{Output}: Generated Image ($32 \times 32 \times 3$)				&
		\textbf{Output}: Real or Fake (Probability)									\\
		
		\hline
		\textbf{DCGAN}	&	\\
		\textbf{Input}: Noise $z \sim p_z$, 100										&	
		\textbf{Input}: Image ($64 \times 64 \times 3$)							\\
		$[$layer 1$]$  Transposed Convolution (4, 4, 1024), stride=1; Batchnorm; \textit{ReLU};	&
		$[$layer 1$]$  Convolution (4, 4, 128), stride=2; \textit{LeakyReLU};		\\
		$[$layer 2$]$  Transposed Convolution (4, 4, 512), stride=2; Batchnorm; \textit{ReLU};	&
		$[$layer 2$]$  Convolution (4, 4, 256), stride=2; Batchnorm; \textit{LeakyReLU};\\
		$[$layer 3$]$  Transposed Convolution (4, 4, 256), stride=2; Batchnorm; \textit{ReLU};	&
		$[$layer 3$]$  Convolution (4, 4, 512), stride=2; Batchnorm; \textit{LeakyReLU};\\
		$[$layer 4$]$  Transposed Convolution (4, 4, 128), stride=2; Batchnorm; \textit{ReLU};	&
		$[$layer 4$]$  Convolution (4, 4, 1024), stride=2; Batchnorm; \textit{LeakyReLU};\\
		$[$layer 5$]$  Transposed Convolution (4, 4, 3), stride=2; \textit{Tanh};	&
		$[$layer 5$]$  Convolution (4, 4, 1), stride=1; \textit{Sigmoid};			\\
		\textbf{Output}: Generated Image ($64 \times 64 \times 3$)				&
		\textbf{Output}: Real or Fake (Probability)									\\
		
		\hline 
		\textbf{MLP}	&    \\ 
		\textbf{Input}: Noise $z \sim p_z$, 2 &  \textbf{Input}: Point ($2 \times 1$) \\
		$[$layer 1$]$  Fully Connected (512); \textit{ReLU}; &  $[$layer 1$]$  Fully Connected (512); \textit{ReLU}; \\
		$[$layer 2$]$  Fully Connected (512); \textit{ReLU}; &  $[$layer 2$]$  Fully Connected (512); \textit{ReLU}; \\
		$[$layer 3$]$  Fully Connected (512); \textit{ReLU}; &  $[$layer 3$]$  Fully Connected (512); \textit{ReLU}; \\
		$[$layer 4$]$  Fully Connected (2); &  $[$layer 4$]$  Fully Connected (1); \textit{Sigmoid}; \\
		\textbf{Output}: Generated Point ($2 \times 1$) &  \textbf{Output}: Real or Fake (Probability) \\
		\hline
	\end{tabular}
\end{table*}

The values of {the} hyperparameters shared with E-GAN {have} the same {values that they have in E-GAN}. Specifically, {the Adam optimizer with {a} learning rate} {$\alpha=2\!\times\!10^{-4}$} {,} $\beta_1=0.5$, {and} $\beta_2=0.999$ {is used}; {the {number of} updating steps of {the} discriminator per iteration} $n_D=3$; {the number of mutations} $n_m=3$; {the mutation batch size} $m=32$; {and the sample size for evaluation} $l=256$. {Additionally, we recommend a population size $\mu = 1$.} {However,} the balance coefficient {for the} fitness quality and diversity (i.e., $\gamma_1$) is an exception. This is due to {how} we redefine the quality and diversity fitness function. We select $\gamma_1=1$ for the synthetic datasets and $\gamma_1=0.05$ for the real-world datasets via grid search.
For IE-GAN{-}specific hyperparameters, we choose $\gamma_2=0.001$, $n_e=5$, and {crossover size} $n_c=1$ for all experiments. {W}e set the crossover batch size $n$ to 256{; it is equal to} $l$. In this paper, we conduct experiments with these values if not otherwise stated.

We use {the} FID \cite{DBLP:conf/nips/HeuselRUNH17} to quantitatively evaluate the generative performance. 
{In fact, the FID is a metric that is widely used in a variety of image synthesis tasks, such as image generation \cite{DBLP:conf/iclr/BrockDS19, DBLP:conf/cvpr/KarrasLA19, DBLP:conf/cvpr/KarrasLAHLA20} and image-to-image translation \cite{DBLP:journals/ijcv/LeeTMHLSY20, zhou2022coutfitgan, zhou2022learning}.}
{The} FID is frequently used as a metric in {studies concerning} GANs, and it is considered superior to other metrics \cite{costa2020neuroevolution}. The lower the FID, the better the quality of the generated images. We randomly generate 50K images to calculate {the} FID {during the test phase}. {Unless otherwise stated}, the experiments in this paper {are} trained for 100K generations.

The experiments are conducted on a single NVIDIA TITAN Xp GPU with 12GB {of} memory and {an} Intel Xeon E5-2620 v2 CPU. The code we implemented in PyTorch is publicly available at \url{https://github.com/AlephZr/IE-GAN}.

\subsection{Mode Collapse}
Learning the 2D Gaussian mixture distribution{s} can visually demonstrate the mode collapse of the model. If the model suffers from mode collapse, the samples it generates will focus on a limited number of modes. {T}his can be directly observed {using} kernel density estimation (KDE) plots.

We compare the complete algorithm of IE-GAN with the baselines{,} including GAN, NS-GAN, LSGAN, and E-GAN. For fairness, all methods utilize the same MLP network architecture. Because of the huge difference between synthetic and real-world datasets, the learning rate and the number of discriminator updates in each iteration are consistent with the source of the network architecture, i.e., {$\alpha=1\!\times\!10^{-4}$} {and} $n_D=1$.

\begin{figure*}[!t]
	\centering
	\begin{tabular}{ccccccc}
		& Target & GAN & NS-GAN & LSGAN & E-GAN & IE-GAN \\
		{\rotatebox{90}{8 Gaussians}} &
		\includegraphics[width=0.13\textwidth]{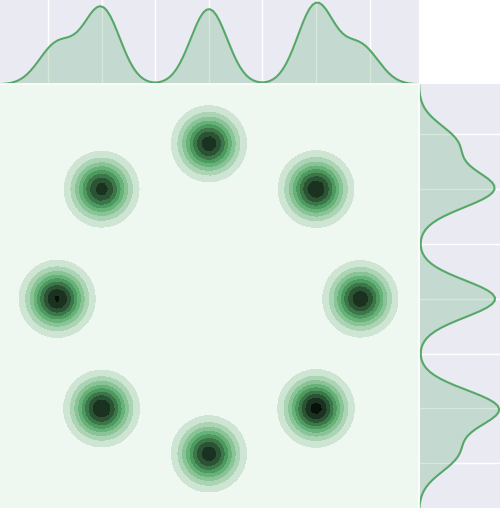} & 
		\includegraphics[width=0.13\textwidth]{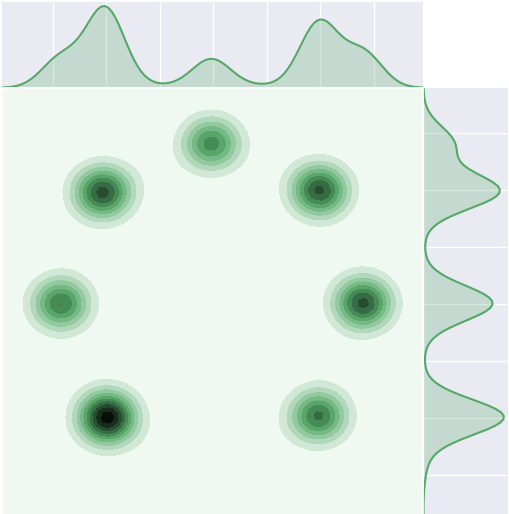} & \includegraphics[width=0.13\textwidth]{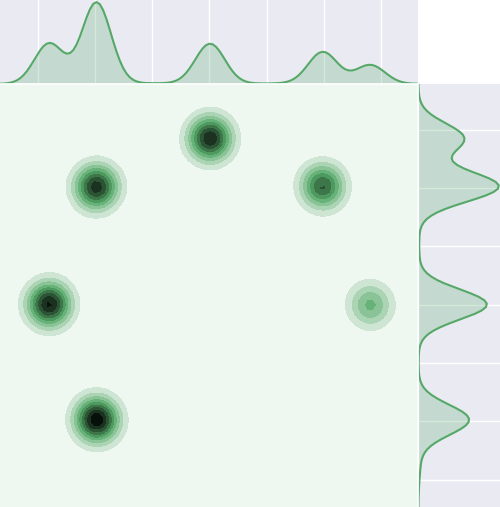} & \includegraphics[width=0.13\textwidth]{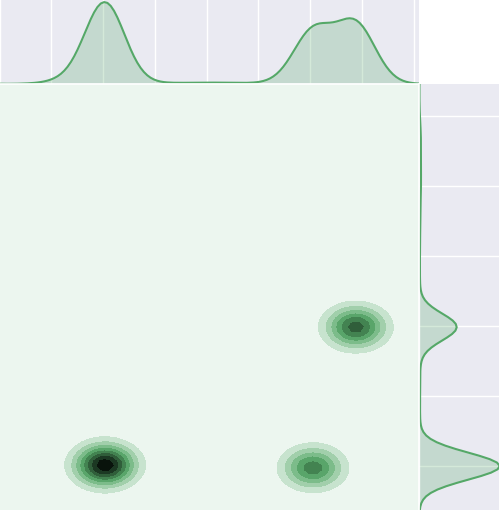} & \includegraphics[width=0.13\textwidth]{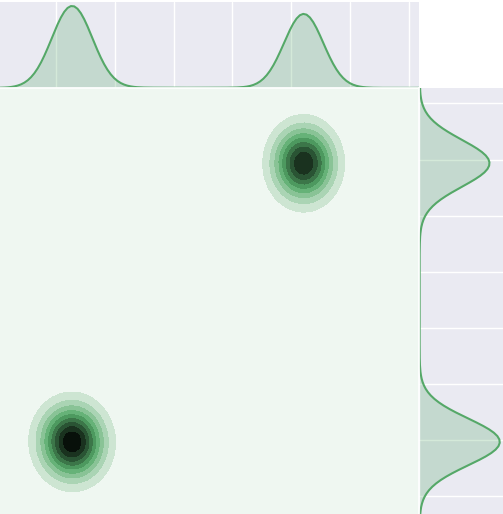} & \includegraphics[width=0.13\textwidth]{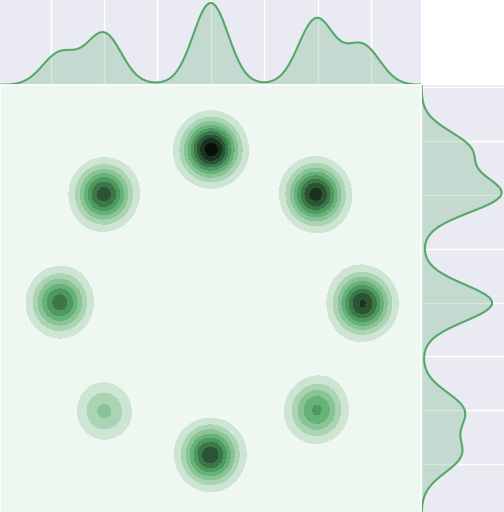} \\
		{\rotatebox{90}{25 Gaussians}} &
		\includegraphics[width=0.13\textwidth]{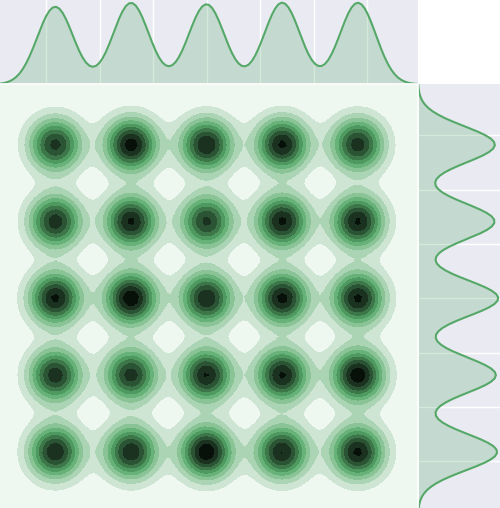} & 
		\includegraphics[width=0.13\textwidth]{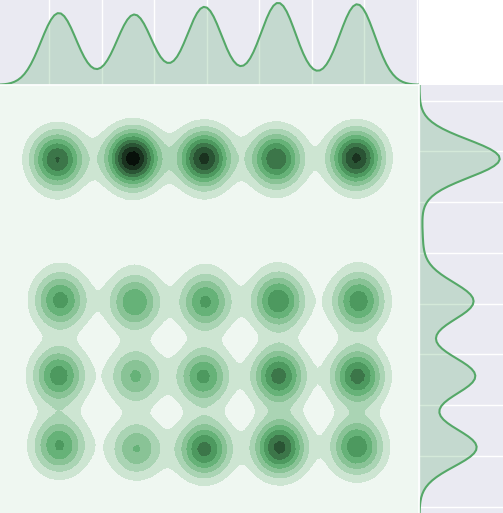} & \includegraphics[width=0.13\textwidth]{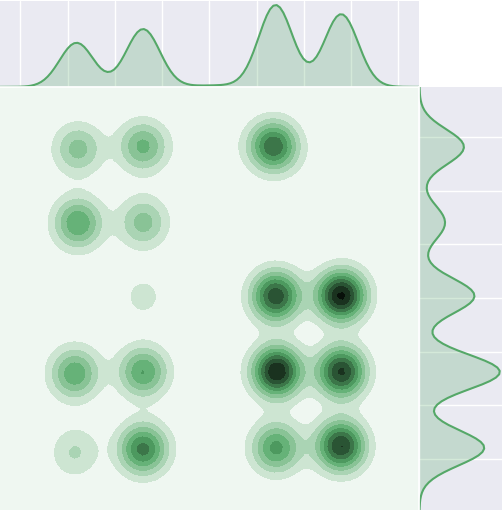} & \includegraphics[width=0.13\textwidth]{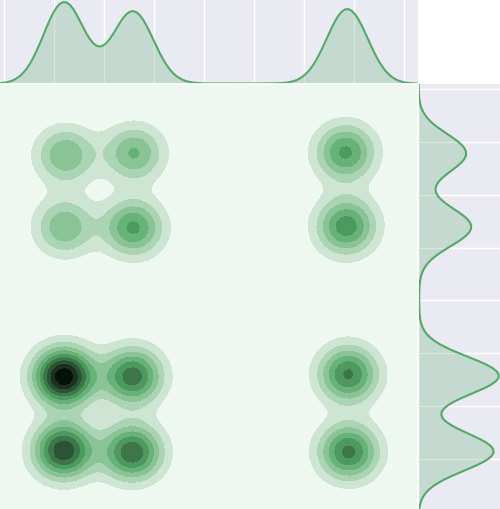} & \includegraphics[width=0.13\textwidth]{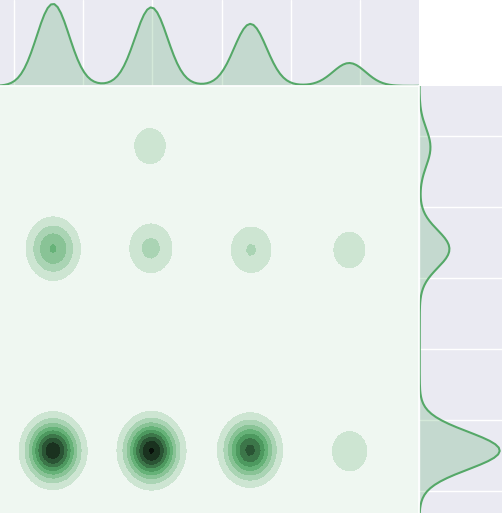} & \includegraphics[width=0.13\textwidth]{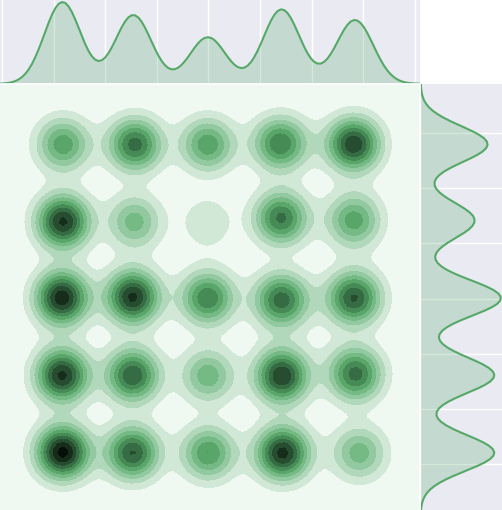} \\						
	\end{tabular}
	\caption{{~KDE plots of the target data and the generated data of various GANs trained on the synthetic dataset. The first row {shows the results for the} 8{-}Gaussian{-}mixture distribution, which is trained {on for} only 30K iterations because of the small number of modes. The second row {shows the results for the} 25{-}Gaussian{-}mixture distribution, {which is} trained {on for} 100K iterations.}} 
	\label{fig:KDEofToy}
	\vspace{-0.5cm}
\end{figure*}

Fig.~\ref{fig:KDEofToy} illustrates the modes captured by different methods. The center plot of each plot is the KDE plot, and the side plots reflect the probability distribution. To adequately represent the generated distribution, {10,240} points are sampled for each plot. We can see that on both synthetic datasets, the baselines show {some} missing {modes}. This is especially true for E-GAN, which appears to be inapplicable to experimental networks. It performs the worst {out} of all {the} methods, and its evolutionary strategy exacerbates mode collapse. {In contrast,} IE-GAN is able to cover all modes (although some modes are weakly covered), which indicates that our evolutionary strategy can effectively suppress mode collapse. This is corroborated by {the} side plots, {which show that} the probability distribution of IE-GAN is closest to {that of} the target data.

\subsection{Generative Performance}
To illustrate the superiority of {the} proposed IE-GAN over {the} baselines, we conduct experiments on the CIFAR-10 dataset \cite{krizhevsky2009learning}. The baselines used for comparison include GAN-Minimax, GAN-Heuristic, and GAN-Least-squares{,} which{,} respectively{,} represent the methods {that use a} single mutation operator. The experimental results obtained by comparing {IE-GAN} with these baselines are used to demonstrate that our approach benefits primarily from the framework rather than {the} network architecture. In order to explore the role of each operator, their settings are identical to {those of} IE-GAN, including $n_D${, the} discriminator objective function{, and the GP}. In addition, we compare {IE-GAN} with E-GAN \cite{DBLP:journals/tec/WangXYT19} to demonstrate the effectiveness of the proposed evolutionary strategy. E-GAN likewise maintains only {one} population {unless otherwise stated}. 

\begin{figure}[!t]
	\centering
	\includegraphics[width=0.45\textwidth]{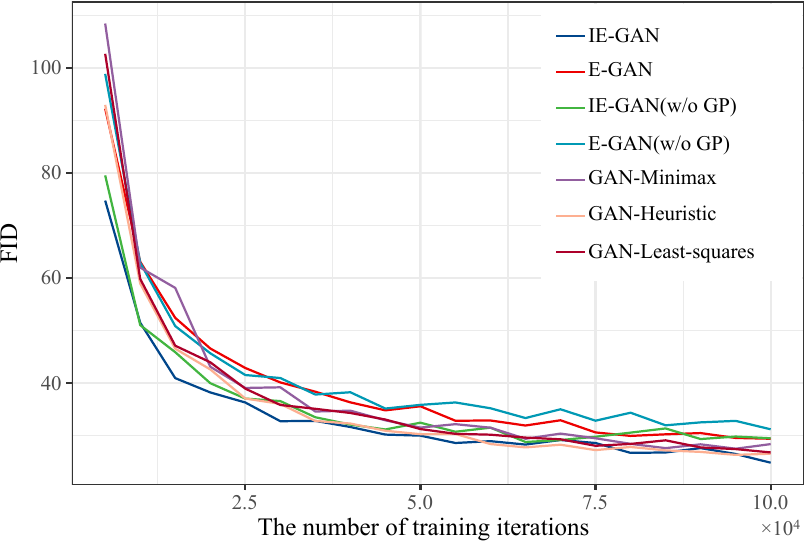}
	\caption{{~{The} FID of different methods on CIFAR-10 over {the} generator iteration{s}. ``w/o GP" indicates that no GP {term} is used.}}
	\label{fig:FIDofMethods}
	\vspace{-0.5cm}
\end{figure}

Fig.~\ref{fig:FIDofMethods} plots {the} FID for {the} training process of different methods {for} the same network architecture. We use {an} experimental setup from the literature \cite{DBLP:journals/tec/WangXYT19}{, along} with codes\footnote{https://github.com/WANG-Chaoyue/EvolutionaryGAN-pytorch} provided by {previous} authors{,} to obtain the results of E-GAN. {The following conclusions can be drawn} from this figure:
\begin{itemize}
	\item IE-GAN {shows a more significant improvement in the generative performance} than E-GAN {due to} the GP term. This is because the GP term acts on the discriminator, which is more deeply involved in IE-GAN. In addition to {participating in} normal gradient learning and in the evaluation{,} as in E-GAN, the discriminator in IE-GAN also takes part in the assessment of the samples during the crossover process. Therefore, IE-GAN can benefit more from discriminator enhancement.
	\item The GP term also accelerates the convergence of IE-GAN, whereas E-GAN {experiences} no such {acceleration}. This is the work of {the} crossover operator. The discriminator is involved in crossover variation, so a better discriminator {makes it possible} to obtain better crossover offspring, which in turn accelerates {the} model convergence.
	\item The crossover operator itself also contributes to {the model} convergence. IE-GAN without {the} GP shows {a} considerable convergence speed, and it is clearly distinguished from methods other than IE-GAN with {the} GP until 25K iterations.
	\item IE-GAN does benefit from multiple operators, and it has the best generative performance and the fastest convergence speed. Most of the time, the fold line representing {IE-GAN} is at the bottom {of the plot}.
	\item The evolutionary strategy of E-GAN works the other way around. Under the same conditions, E-GAN is worse than any of the single{-}mutation{-}operator methods. Its generative performance is only similar to that of IE-GAN without {the} GP term.
\end{itemize}

For comparison and analysis, the FID {values} of different methods including various populations of E-GAN are listed in Table~\ref{tab:FIDofMethods}, where the data marked with $\dagger$ are quoted from \cite{DBLP:journals/tec/WangXYT19}. ``Final FID" in the table denotes the FID after 100K iterations. The best value is bolded. We reproduced the multipopulation experiments of E-GAN, but perhaps because of the different experimental environments, the final results are different from those in \cite{DBLP:journals/tec/WangXYT19}. The results of our experiment show {the following}:
\begin{itemize}
	\item The generative performance of E-GAN can be improved by increasing the population, but {this improvement} is insignificant. As the population size increases from one to eight, the FID decreases by less than one, and the number of maintained generative networks increases from 3 to 24, which increases the cost.
	\item This improvement of E-GAN is also unstable. When $\mu=4$, the experimental results are worse than {they are for} $\mu=1$.
	\item E-GAN does not benefit from multiple operators. The population brings such a small improvement that the performance {at} $\mu=8$ is likely to be close to the limit that E-GAN can achieve. Even so, E-GAN is still inferior to GAN-Minimax, which is the worst single{-}mutation{-}operator method. The results show that, due to the unreasonable evaluation function, E-GAN performs worse than any {other} reference {baseline}.
\end{itemize}

\begin{table}[!t]
	\caption{Comparison of the Results of Different Methods and Various Populations in Terms of the FID}
	\label{tab:FIDofMethods}
	\centering
	\begin{tabular}{lc}
		\toprule
		\makecell[l]{\textbf{Method}} & {\textbf{Final FID}} \\
		\midrule
		{GAN-Minimax} & {$28.36$} \\
		{GAN-Heuristic} & {$26.53$} \\ 
		{GAN-Least-squares} & {$26.77$} \\ 
		\midrule
		{{E-GAN ($\mu = 1$, w/o GP)}$\dagger$} & {$36.2$} \\ 
		{E-GAN ($\mu = 1$)$\dagger$} & {$33.2$} \\
		{E-GAN ($\mu = 2$)$\dagger$} & {$31.6$} \\ 
		{E-GAN ($\mu = 4$)$\dagger$} & {$29.8$} \\ 
		{E-GAN ($\mu = 8$)$\dagger$} & {$27.3$} \\ 
		{{E-GAN ($\mu = 1$, w/o GP)}} & {$31.18$} \\ 
		{E-GAN ($\mu = 1$)} & {$29.38$} \\
		{E-GAN ($\mu = 2$)} & {$28.84$} \\ 
		{E-GAN ($\mu = 4$)} & {$30.46$} \\ 
		{E-GAN ($\mu = 8$)} & {$28.45$} \\ 
		\midrule
		{{IE-GAN (Ours) ($\mu = 1$, w/o GP)}} & {$29.47$} \\ 
		{IE-GAN (Ours) ($\mu = 1$)} & $\textbf{24.82}$ \\ 
		\bottomrule
	\end{tabular}
	\vspace{-0.5cm}
\end{table}

{Finally, IE-GAN is compared with seven typical evolutionary GANs to {demonstrate} the advantages of its evolutionary strategies. {The} FIDs of all methods {for} the CIFAR-10 dataset are reported in Table~\ref{tab:FIDofEGANs}. ``30K FID" in the table denotes the FID after 30K iterations to show the convergence speeds of the methods. Evolution-GAN \cite{DBLP:conf/gecco/GarciarenaSM18}, COEGAN \cite{DBLP:conf/gecco/Costa0CM19}, and QD-COEGAN \cite{DBLP:conf/gecco/Costa0CM20} do not record 30K FIDs because of the differences in {their} training strategies. The best results are {underlined}. $\mu$ and $\nu$ in Table~\ref{tab:FIDofEGANs} represent the {generator and discriminator} population sizes maintained by the evolutionary methods, respectively. The experimental code {for each method} is obtained from the corresponding study, and the configuration refers to the source description.}

\begin{table}[!t]
	\caption{{Comparison of the Results of Different Evolutionary GANs in Terms of the FID}}
	\label{tab:FIDofEGANs}
	\centering
	\begin{threeparttable}
		\begin{tabular}{lcc}
			\toprule
			\makecell[l]{\textbf{Method}} & {\textbf{30K FID}} & {\textbf{Final FID}} \\
			\midrule
			{E-GAN ($\mu = 1$) {\cite{DBLP:journals/tec/WangXYT19}}} & {$40.10$} & {$29.38$} \\
			{MO-EGAN ($\mu = 8$) {\cite{DBLP:conf/gecco/BaiolettiCBP20}}} & {$36.26$} & {$29.93$} \\ 
			{SMO-EGAN ($\mu = 8$)  {\cite{DBLP:conf/cec/BaiolettiBPC21}}} & {$37.93$} & {$29.46$} \\ 
			{CDE-GAN ($\mu = 1, \nu = 2$)  {\cite{DBLP:journals/tec/ChenWXYPCD21}}} & {$68.24$} & {$27.57$} \\ 
			\midrule
			{Evolution-GAN ($\mu = 20, \nu = 20$) {\cite{DBLP:conf/gecco/GarciarenaSM18}}} & {$-$} & {$-$} \\ 
			{COEGAN ($\mu = 10, \nu = 10$) {\cite{DBLP:conf/gecco/Costa0CM19}}} & {$-$} & {$85.92$} \\
			{QD-COEGAN ($\mu = 5, \nu = 5$) {\cite{DBLP:conf/gecco/Costa0CM20}}} & {$-$} & {$169.40$} \\ 
			\midrule
			{IE-GAN {(Ours)} ($\mu = 1$)} & {$\underline{\textbf{32.72}}$\tnote{12}} & {$\underline{24.82}$} \\ 
			{\makecell[l]{CDE-GAN (IE-GAN framework) \\ ($\mu = 1, \nu = 2$)}} & {$51.79$} & {$\textbf{24.33}$} \\ 
			\bottomrule
		\end{tabular}
		\begin{tablenotes}
			\footnotesize
			\item[1] The best results for the unmodified methods are underlined.
			\item[2] The best results out of all methods are shown in bold.
		\end{tablenotes}
	\end{threeparttable}
	\vspace{-0.5cm}
\end{table}

{The experiments show that IE-GAN outperforms all baselines. The first group {in} Table~\ref{tab:FIDofEGANs}, the learning-based evolutionary GANs, all perform poorly compared to our IE-GAN. It is worth noting that the {default numbers of iterations} of MO-EGAN \cite{DBLP:conf/gecco/BaiolettiCBP20}, SMO-EGAN \cite{DBLP:conf/cec/BaiolettiBPC21} and CDE-GAN \cite{DBLP:journals/tec/ChenWXYPCD21} are larger than the 100K iterations used by IE-GAN{;} MO-EGAN and SMO-EGAN run for 100 epochs (about 150K iterations), and CDE-GAN runs for 150K iterations. The second group of methods in Table~\ref{tab:FIDofEGANs} randomly vary the coding genes. Experiments demonstrate that an inherently unstable system like {a} GAN should not introduce excessive randomness when dealing with complex tasks. The disorderly direction of variation forces the population to maintain a large size, {which results in} greater resource {use}. {Additionally,} this disorderly exploration is too inefficient for complex task spaces to be competitive. In particular, Evolution-GAN \cite{DBLP:conf/gecco/GarciarenaSM18}, which has the strongest randomness among all methods and has a network structure that only supports fully connected layers, cannot achieve meaningful results on the CIFAR-10 dataset.}
{Meanwhile, the framework of IE-GAN is applied to CDE-GAN to verify its generalizability. The experimental results are recorded in Table~\ref{tab:FIDofEGANs}. The best results are shown in bold. It can be seen that IE-GAN effectively improves the convergence speed and generation performance of CDE-GAN. Moreover, none of the hyperparameters involved in the experiments are tuned. This proves that IE-GAN has good robustness{; it did not achieve good results} because of carefully tuned hyperparameters.}

\subsection{{Time Cost}}
\begin{figure}[!t]
	\centering
	\includegraphics[width=0.485\textwidth]{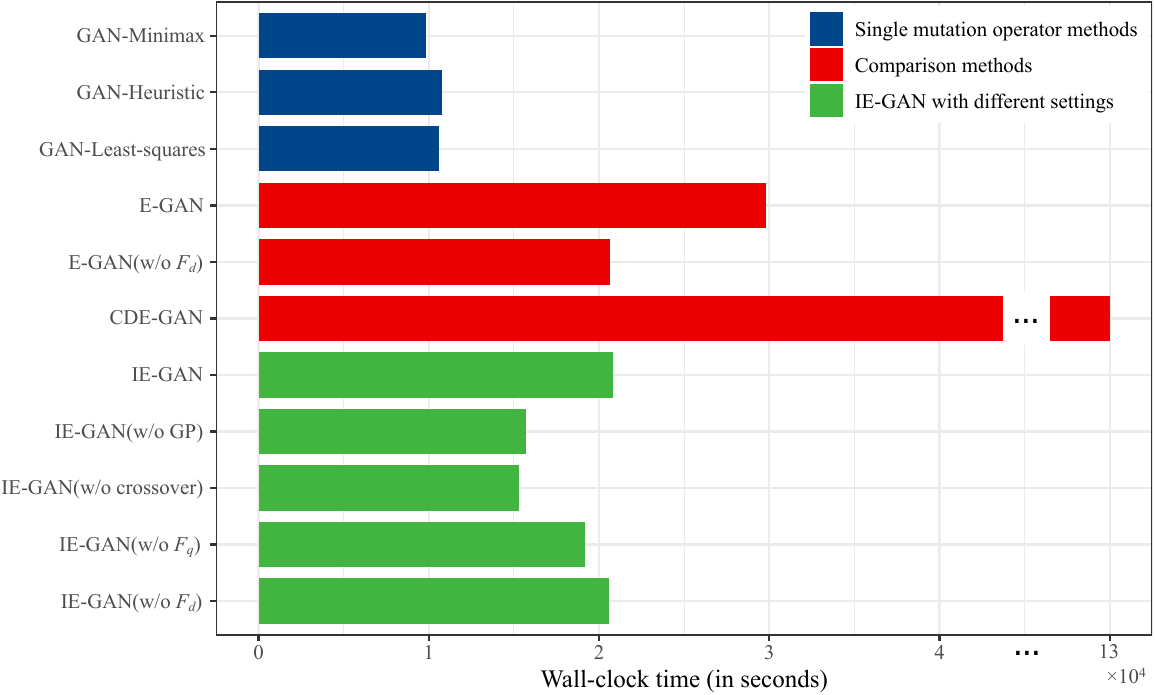}
	\caption{{~Wall-clock time of different methods {and different settings}.}}
	\label{fig:Wall-clockofMethods}
	\vspace{-0.5cm}
\end{figure}

{The r}untime is also an important criterion for evaluating method usability. Fig.~\ref{fig:Wall-clockofMethods} shows {the} wall-clock time {required to train} different methods {for} 100K iterations on the CIFAR-10 dataset, and the length of {each} bar reflects {the} time cost of the corresponding method.

GAN-Minimax, GAN-Heuristic, and GAN-Least-squares maintain only one generator network and no additional evaluation of the generators is required, so the time cost {of} training {these networks} is {similar and less than that of IE-GAN}. 
{Among the evolutionary GANs, the proposed IE-GAN shows a significant advantage in terms of the time cost. IE-GAN maintains four generative networks, while {requiring} about twice as much time as the single{-}mutation{-}operator methods. The time cost of E-GAN is obviously higher than that of IE-GAN, although it does not produce crossover offspring. While CDE-GAN maintains only three more networks than IE-GAN, its time cost is more than six times that of IE-GAN.}

{On the one hand, the dominance of IE-GAN is attributed to the relatively simple diversity fitness function. The IE-GAN experiments with different settings in Fig.~\ref{fig:Wall-clockofMethods} demonstrate the low consumption of $F^{IE-GAN}_d$. The reduction in {the} wall-clock time {caused} by removing $F_d$ from IE-GAN is very small. The calculation {time} of $F^{IE-GAN}_d$ is the lowest {out of} all {the} components, and {it} has little effect on {the} time {cost}. In contrast, $F^{E-GAN}_d$ {requires} considerable calculation{s using} the discriminator gradient information. As shown in Fig.~\ref{fig:Wall-clockofMethods}, removing $F_d$ can effectively accelerate the training of E-GAN. On the other hand, the wall-clock time advantage of IE-GAN {benefits} from the streamlined framework structure. The reuse of the samples generated in the evaluation phase during crossover offspring learning minimizes the {time needed by the} training process. CDE-GAN {requires more time due to} the coevolution of the two populations. CDE-GAN maintains about twice the number of networks as E-GAN, but the wall-clock time is more than four times that of E-GAN. This suggests that it is inefficient not only because it inherits $F_d$ from E-GAN but also due to its competitive strategy. The two populations of CDE-GAN must score each other, which greatly increases the cost of evaluation. In summary, the efficient $F_d$ and the brief training process together {result in} the low time cost of IE-GAN.}

\subsection{Ablation Study}
\begin{figure*}[!t]
	\centering
	\setlength{\tabcolsep}{4.5mm}{
		\begin{tabular}{cccccc}
			& Target & \makecell[c]{IE-GAN\\(w/o $F_d$)} & \makecell[c]{IE-GAN\\(w/o crossover)} & \makecell[c]{IE-GAN\\(neither)} & IE-GAN \\
			{\rotatebox{90}{8 Gaussians}} &
			\includegraphics[width=0.13\textwidth]{img/samples/8_true_dist.png} & 
			\includegraphics[width=0.13\textwidth]{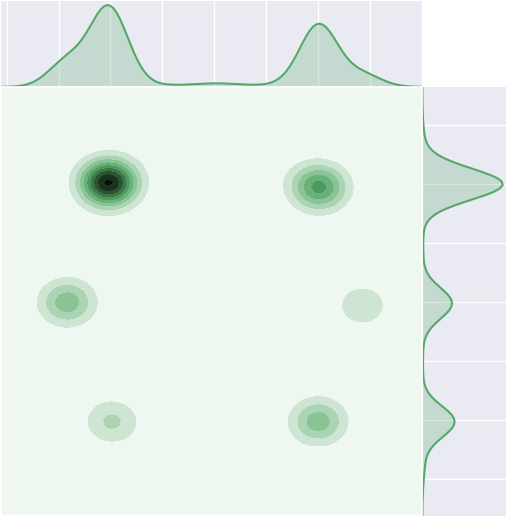} & \includegraphics[width=0.13\textwidth]{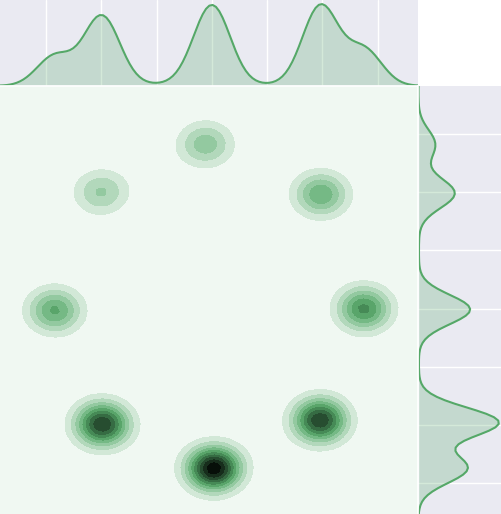} & \includegraphics[width=0.13\textwidth]{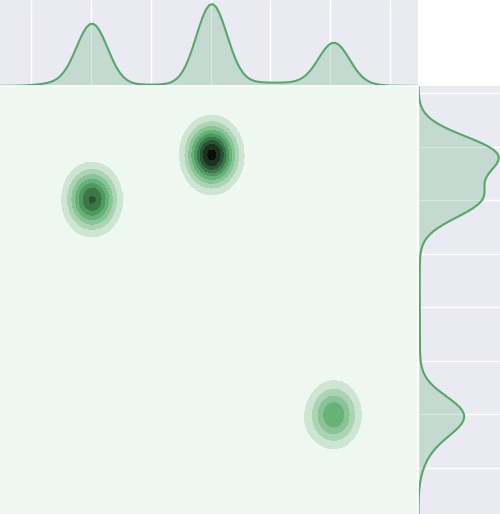} &  \includegraphics[width=0.13\textwidth]{img/samples/IE-GAN_kde_frame_30.png} \\
			{\rotatebox{90}{25 Gaussians}} &
			\includegraphics[width=0.13\textwidth]{img/samples/25_true_dist.png} & 
			\includegraphics[width=0.13\textwidth]{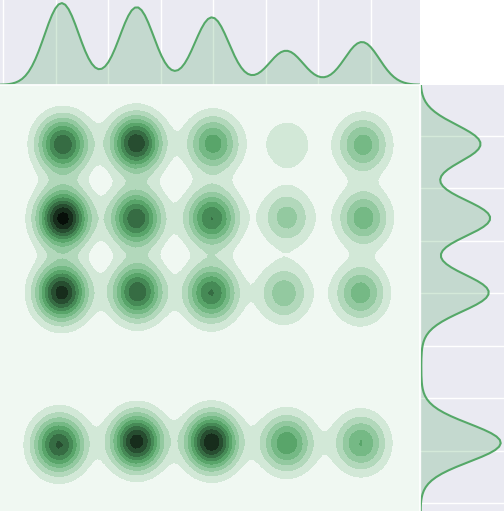} & \includegraphics[width=0.13\textwidth]{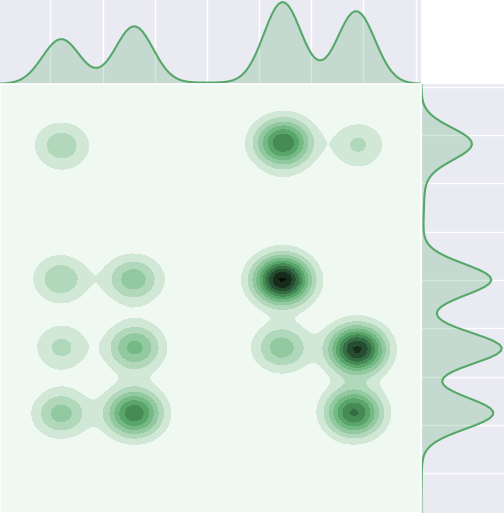} & \includegraphics[width=0.13\textwidth]{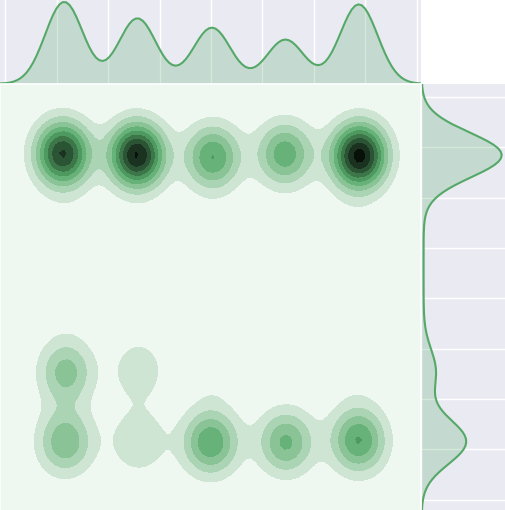} &  \includegraphics[width=0.13\textwidth]{img/samples/IE-GAN_kde_25gaussian_100.png} \\						
		\end{tabular}
	}
	\caption{{~KDE plots for {the} ablation study. The first row {shows the results for the 8-}Gaussian{-}mixture distribution, {which is} trained {on} for 30K iterations. The second row {shows the results for the} 25{-}Gaussian{-}mixture distribution, {which is} trained {on} for 100K iterations. ``w/o $F_d$" indicates that IE-GAN does not use the diversity fitness function $F^{IE-GAN}_d$. ``w/o crossover" indicates that IE-GAN does not apply crossover variation. ``neither" indicates that IE-GAN {uses} neither {the diversity fitness function nor crossover variation}.}}
	\label{fig:KDEofToyAblation}
	\vspace{-0.5cm}
\end{figure*}

\begin{figure}[!t]
	\centering
	\includegraphics[width=0.45\textwidth]{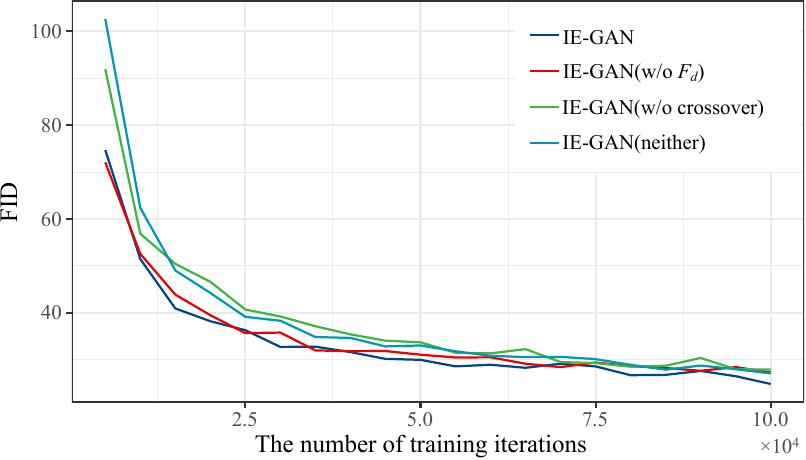}
	\caption{{~{The} FID for {the} ablation study. ``w/o $F_d$" indicates that IE-GAN does not use the diversity fitness function $F^{IE-GAN}_d$. ``w/o crossover" indicates that IE-GAN does not apply crossover variation. ``neither" indicates that IE-GAN {used} neither {the diversity fitness function nor crossover variation}.}}
	\label{fig:FIDofAblation}
	\vspace{-0.5cm}
\end{figure}

We compare IE-GAN with its ablation variants. Fig.~\ref{fig:KDEofToyAblation} shows examples {of IE-GAN and its ablation on} the synthetic datasets. Eliminating $F^{IE-GAN}_d$ or {the} crossover variation decreases {the} performance{,} and eliminating both leads to more serious consequences. 
This means that both $F^{IE-GAN}_d$ and crossover support generative diversity.
{A s}imilar conclusion can be obtained on the CIFAR-10 dataset. Fig.~\ref{fig:FIDofAblation} shows that both items are critical to our method. In addition, the crossover operator exhibits its acceleration effect.

\subsection{Hyperparameter Analysis}
We use the same discriminator objective function as E-GAN.
There are two more hyperparameters that are closely related to the performance of IE-GAN, namely the balance coefficient{s} of {the} fitness function ($\gamma_1$) and {the} sample evaluation function ($\gamma_2$). Since they are used to balance quality and diversity measurements, they guide the selection of generators and samples, which affects the performance of IE-GAN. Meanwhile, the targets of {the} knowledge distillation of {the} crossover operator, the number of parents imitated by crossover offspring, the crossover batch size $n$, and the {number} of sample comparisons $n_e$ all affect the effectiveness of IE-GAN. In this {sub}section, we analyze their effects one by one through experiments on CIFAR-10.

\subsubsection{$\gamma_1$ Correlation}
{The fitness function represents the pressure of the environment on individuals. The effects of the two types of metrics, quality and diversity, in the individual fitness measure are balanced by the coefficient $\gamma_1$. This coefficient directly affects the selection of the population by the environment and must therefore be {chosen} carefully.}

In order to guarantee confidence, we choose a larger population to ensure that there are enough individuals for each evaluation. In this way, the effect of {the} balance coefficient $\gamma_1$ can be amplified so that the performance{s} of the generators {are} sufficiently differentiated. In this experiment, the population size is set to $\mu = 4$. To eliminate the interference of the crossover operator and related hyperparameters, the crossover operator is not enabled in this experiment.
We perform a grid search to find {an} appropriate value {of} $\gamma_1$. Table~\ref{tab:gamma1} lists the performance measurements of the trained generators{;} the balance coefficient{s} $\gamma_1 \in \{0, 0.001, 0.01, 0.05, 0.1, 0.5, 1\}$ are {used in} the experiment. When $\gamma_1=0$, $F^{IE-GAN}_d$ is not enabled. ``Best FID" in the table represents the best FID of the method during the training process. The best value for each item is bolded.

\begin{table}[!t]
	\caption{{Comparison of the} Results of Various Fitness Balance Coefficient{s} in Terms of {the} FID (w/o Crossover)}
	\label{tab:gamma1}
	\centering
	\begin{tabular}{lcc}
		\toprule
		\makecell[l]{\textbf{Method}} & {\textbf{Final FID}} & {\textbf{Best FID}} \\
		\midrule
		{IE-GAN ($\gamma_1 = 0$)} & {$27.38$} & {$27.38$} \\
		{IE-GAN ($\gamma_1 = 0.001$)} & {$28.11$} & {$27.59$} \\ 
		{IE-GAN ($\gamma_1 = 0.01$)} & {$27.79$} & {$27.79$} \\ 				
		{IE-GAN ($\gamma_1 = 0.05$)} & {$\textbf{27.15}$} & {$\textbf{26.74}$} \\ 				
		{IE-GAN ($\gamma_1 = 0.1$)} & {$27.21$} & {$27.21$} \\ 				
		{IE-GAN ($\gamma_1 = 0.5$)} & {$27.86$} & {$27.56$} \\ 
		{IE-GAN ($\gamma_1 = 1$)} & {$27.73$} & {$27.73$} \\ 
		\bottomrule
	\end{tabular}
	\vspace{-0.5cm}
\end{table}

Whether {the} final FID or {the} best FID {is considered}, these values generally decrease first and then increase as $\gamma_1$ increases. The exception is $\gamma_1=0$, which {provides a} better score than any of the {values} outside $\gamma_1\in\{0.05, 0.1\}$. This suggests that an inappropriate additional term is more likely to be a source of interference.
The results show that the performance is correlated with $\gamma_1$, which leads to poorer consequences {if} it is too large or too small. This phenomenon occurs in experiments with different population sizes. We select $\gamma_1 = 0.05$ for real-world image datasets based on observations.

\subsubsection{$\gamma_2$ Correlation}

{The balance coefficient $\gamma_2$ of the sample evaluation function $E$ affects {the} competitiveness of {the} crossover offspring by controlling the learning objects. Appropriate learning objects mean that the crossover offspring can collect more of the superior {characteristics} of the parents rather than their defects. The crossover operator that depends on $\gamma_2$ is expected to generate dominant offspring, while the fitness function that includes $\gamma_1$ represents the desired evolutionary direction. Although the two behave similarly, they should not be roughly equal.}

{In addition to the sample evaluation function, the knowledge distillation targets also affect the position of {the} crossover offspring in the population. Although softer targets are preferred in knowledge distillation \cite{DBLP:journals/corr/HintonVD15}, as they minimizes the information loss, their effectiveness should be experimentally verified in different algorithmic settings.}
{Likewise, the number of crossover parents can affect system outcomes through the generated offspring.}
In principle, our crossover operator allows the offspring to learn from a set of networks rather than a limited set of two parents. 
{If the crossover offspring is precise enough to absorb the experience of its parents, it can learn more from more trained networks.}

Since all three hyperparameters ($\gamma_2$, knowledge distillation targets, and {the nubmer of} crossover parents) are related to the crossover operator and are tightly correlated, they are considered together here. We perform a grid search for $\gamma_2 \in \{0, 0.0001, 0.001, 0.01, 0.1, 1\}$ in a total of four contexts combined with different knowledge distillation targets and number{s} of crossover parents. We set the size of {the} population $\mu=1$ and the crossover {size} $n_c=1$ in this experiment. 
The performance measurements are reported in Table~\ref{tab:gamma2|cpop|target}. The best {result in} each context is bolded and the best {result out} of all contexts is additionally underlined. 

\begin{table*}[!t]
	\caption{{Comparison of the} Results of {Different} Knowledge Distillation Targets, {Numbers of} Crossover Parents, and Sample Evaluation Balance Coefficient{s} in Terms of {the} FID}
	\label{tab:gamma2|cpop|target}
	\centering
	\begin{threeparttable}
		\begin{tabular}{c|lcc|lcc}
			\hline
			\multicolumn{1}{l|}{}                 & \multicolumn{3}{c|}{\textbf{Soft Targets}}          & \multicolumn{3}{c}{\textbf{Hard Targets}}          \\ \hline
			\multirow{7}{*}{\rotatebox{90}{\textbf{{Two} Parents}}}   & \multicolumn{1}{l}{\textbf{Method}} & \textbf{Final FID} & \textbf{Best FID} & \multicolumn{1}{l}{\textbf{Method}} & \textbf{Final FID} & \textbf{Best FID} \\ \cline{2-7} 
			& IE-GAN ($\gamma_2 = 0$) & $25.79$ & $25.74$ & IE-GAN ($\gamma_2 = 0$) & $25.65$ & $25.65$ \\
			& IE-GAN ($\gamma_2 = 0.0001$) & $26.11$ & $26.11$ & IE-GAN ($\gamma_2 = 0.0001$) & $26.17$ & $26.17$ \\
			& IE-GAN ($\gamma_2 = 0.001$)  & $\underline{\textbf{24.82}}\tnote{12}$ & $\underline{\textbf{24.82}}$ & IE-GAN ($\gamma_2 = 0.001$)  & $\textbf{25.63}$ & $\textbf{25.63}$ \\
			& IE-GAN ($\gamma_2 = 0.01$)   & $27.49$  & $25.70$ & IE-GAN ($\gamma_2 = 0.01$)   & $26.52$ & $26.52$ \\
			& IE-GAN ($\gamma_2 = 0.1$)    & $26.27$ & $25.82$ & IE-GAN ($\gamma_2 = 0.1$)    & $25.84$ & $25.84$ \\
			& IE-GAN ($\gamma_2 = 1$)      & $26.02$ & $26.02$ & IE-GAN ($\gamma_2 = 1$)      & $27.08$ & $26.34$ \\ \hline
			\multirow{7}{*}{\rotatebox{90}{\textbf{{All} Parents}}} & \multicolumn{1}{l}{\textbf{Method}} & \textbf{Final FID} & \textbf{Best FID} & \multicolumn{1}{l}{\textbf{Method}} & \textbf{Final FID} & \textbf{Best FID} \\ \cline{2-7} 
			& IE-GAN ($\gamma_2 = 0$) & $26.71$ & $26.23$ & IE-GAN ($\gamma_2 = 0$) & $26.63$ & $26.37$ \\
			& IE-GAN ($\gamma_2 = 0.0001$) & $26.13$ & $26.13$ & IE-GAN ($\gamma_2 = 0.0001$) & $\textbf{26.19}$ & $26.19$ \\
			& IE-GAN ($\gamma_2 = 0.001$)  & $\textbf{25.81}$ & $\textbf{25.81}$ & IE-GAN ($\gamma_2 = 0.001$)  & $26.64$ & $26.64$ \\
			& IE-GAN ($\gamma_2 = 0.01$)   & $26.36$ & $26.36$ & IE-GAN ($\gamma_2 = 0.01$)   & $26.21$ & $\textbf{25.94}$ \\
			& IE-GAN ($\gamma_2 = 0.1$)    & $26.70$  & $26.70$ & IE-GAN ($\gamma_2 = 0.1$)    & $27.82$ & $26.45$ \\
			& IE-GAN ($\gamma_2 = 1$)      & $26.31$ & $26.07$ & IE-GAN ($\gamma_2 = 1$)      & $26.79$ & $26.79$ \\ \hline
		\end{tabular}
		\begin{tablenotes}
			\footnotesize
			\item[1] The best results in each context are shown in bold.
			\item[2] The best results out of all contexts are underlined.
		\end{tablenotes}
	\end{threeparttable}
	\vspace{-0.5cm}
\end{table*}

Within each context, equation \eqref{eq:Ed} is counterproductive when $\gamma_2$ is inappropriate, and $E_{d}$ gradually takes on a positive effect as $\gamma_2$ tends to{ward} 0.001 in general. {T}his is not true in the context of hard targets {when} all mutation offspring {are used} as crossover parents{; in this case, the} FID is abnormal at $\gamma_2=0.001$. We believe that {this} outlier is due to the randomness of the experiment. 
In summary, we set $\gamma_2$ to 0.001. Experiments show that this value is applicable to both real-world image datasets and synthetic datasets.
{In addition,} the comparison between contexts can {justify} the choice of the other two hyperparameters. The crossover operator with soft targets is more promising than that with hard targets as expected. However, more crossover parents do not bring greater advantages. With both soft and hard targets, the results of the {two}-parent methods are mostly better than the methods with more parents, in terms of {both the} final FID and {the} best FID. Therefore, our crossover operator employs soft targets in knowledge distillation{,} and {each offspring receives} experience from only one pair of parents.

We presume that the reason for the negative impact of more crossover parents is that $E$ does not fit reality well enough. The fixed balance coefficient $\gamma_2$ constrains the efficiency of $E$. Intuitively, as Fig.~\ref{fig:Fd&Samples} illustrates, quality evaluation is more important in the early stage of training, when normal training can boost {the} diversity at a high rate; {by contrast,} diversity evaluation is more important in the later stage of training, when {the} quality is difficult to improve{,} and diversity loss should be avoided as much as possible.
{T}he effectiveness of methods can be ensured by limiting the influence of this underfitting function on the training process, so a pair of crossover parents is more favorable for our work. 

\subsubsection{Other}
{In evolutionary computations, it is usually considered that when resources are sufficient, the more individuals {there} are {in the population,} the more the evolutionary potential of the population can be exploited. Namely, the population size $\mu$ and the number of crossover offspring $n_c$ should be positively correlated with the system performance. However, in practice, the fitness function representing the evolutionary pressure does not always fit the needs {of the system}.}
Based on the reality that $\gamma_1$ suffers from the same problem as $\gamma_2$, the number of evaluated individuals should be negatively correlated with the generative performance. 
{Therefore, we enlarge the population and increase the number of crossover offspring to verify the assumption in the previous subsection.}

\begin{figure}[!t]
	\centering
	\includegraphics[width=0.45\textwidth]{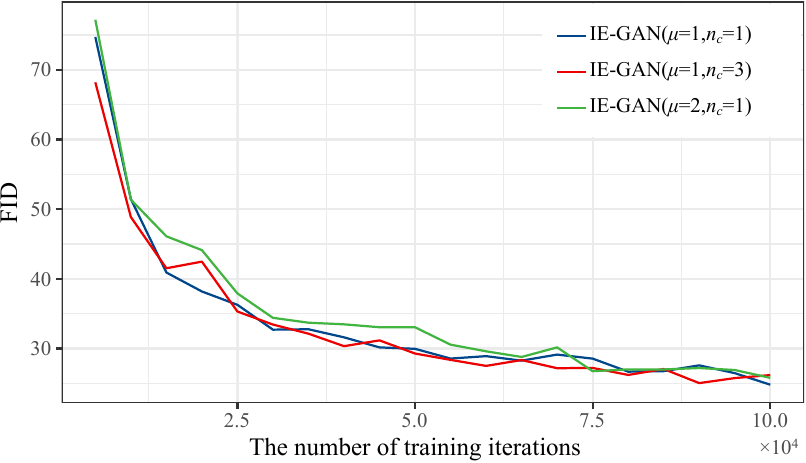}
	\caption{{~{The} FID of different IE-GANs with various crossover size{s} {(}$n_c\in\{1,3\}${)} and population size{s} {(}$\mu\in\{1,2\}${)}.}}
	\label{fig:InfofNum}
	\vspace{-0.5cm}
\end{figure}

As shown in Fig.~\ref{fig:InfofNum}, {a} larger population brings {mostly} negative effects: slow convergence, unstable training, and {a} poor FID. This confirms our {assumption} that in the face of a larger population, $F^{IE-GAN}$ seems to be a little weak. Therefore, we recommend that the population maintain only one size.
{M}ore crossover offspring perform relatively well{;} IE-GAN with $n_c=3$ reflects {the} advantage {of} having a similar number of generative networks as IE-GAN with $\mu=2$. {This} accelerates the convergence{,} and {this IE-GAN's} best FID is comparable to that of a single crossover offspring. This is perhaps due to the excellent crossover operator itself, which improves the fault tolerance of {the} evaluation. {However,} probably because of the fast convergence, it appears to {experience} a bit {of} {overfitting} in the late stage. {Additionally,} it does not observably improve {the} FID. Considering the increase in {the} computing overhead due to more generative networks, we choose $n_c=1$.

\begin{figure}[!t]
	\centering
	\includegraphics[width=0.45\textwidth]{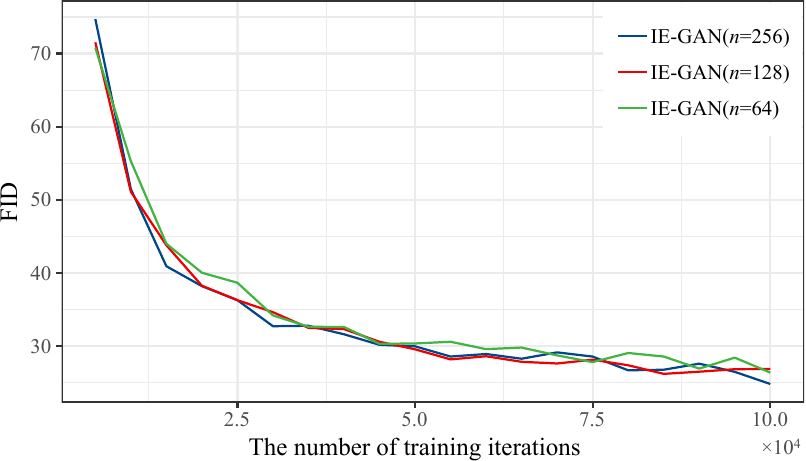}
	\caption{{~{The} FID of different IE-GANs with various crossover batch size{s} {(}$n\in\{256,128,64\}${)}.}}
	\label{fig:CrossBS}
\end{figure}

{In addition to the total number of individuals evaluated, the survival performance of individuals can also interfere with the fitness of the population.}
The crossover operator transfers the experience of parental networks {using} knowledge distillation, and {the} crossover batch size $n$ determines the {amount} of experience that offspring can {access}. Logically, the larger $n$ is, the {more} the offspring understands the parental strategy and the better it inherits the superior {characteristics}. We conduct experiments on this {value} and record the results in Fig.~\ref{fig:CrossBS}.
As expected, the experiments show that the competitiveness of {the} method is positively correlated with $n$ within a certain range. Along with the increase of $n$, IE-GAN {improves} in {terms of the} convergence speed, stability, and generative performance. The crossover batch size $n$ is set to 256 in this paper{; this} is consistent with the number of samples {used to calculate the} fitness {value}. As mentioned earlier, it is an established {practice} to evaluate the fitness of mutation offspring before crossover variation. Fully reusing the {evaluation results}, including {the} generated samples and single-sample evaluation{s}, can {juggle} performance and time cost.

{The factors affecting {the} system performance also include $n_e$, which} represents how many samples each sample should be compared with when scoring {the} diversity. 
{The diversity fitness value, as one of the references for the fitness value, is involved in the selection of surviving individuals for population and learning objects for crossover. Its accuracy is crucial for the effectiveness of the system.}
{The experimental results {concerning} the value of $n_e$ are listed in Table~\ref{tab:ComSize}. Experiments show that if {this} value is too large{, this leads} to repeated comparison{s}, thus wasting resources while hindering {the} reliability {of the fitness value}. However, if the value of $n_e$ is too small, the samples used to calculate the diversity fitness value} cannot be effectively distinguished and $F^{IE-GAN}_d$ does not work. {According to the experiments,} we think {that} $n_e=5$ is appropriate.

\begin{table}[!t]
	\caption{{Comparison of the} Results of Various Sample Number{s}}
	\label{tab:ComSize}
	\centering
	\begin{tabular}{lcc}
		\toprule
		\makecell[l]{\textbf{Method}} & {\textbf{Final FID}} & {\textbf{Best FID}} \\
		\midrule
		{IE-GAN ($n_e = 1$)} & {$26.66$} & {$26.66$} \\
		{IE-GAN ($n_e = 3$)} & {$26.07$} & {$25.54$} \\
		{IE-GAN ($n_e = 5$)} & {$\textbf{24.82}$} & {$\textbf{24.82}$} \\
		{IE-GAN ($n_e = 30$)} & {$26.70$} & {$26.20$} \\
		{IE-GAN ($n_e = 256$)} & {$27.15$} & {$26.58$} \\
		\bottomrule
	\end{tabular}
	\vspace{-0.5cm}
\end{table}

\vspace{-0.2cm}
\subsection{Diversity Fitness Function Analysis}
\begin{figure*}[!t]
	\centering
	\subfloat[20K iterations]{
		\includegraphics[width=0.45\textwidth]{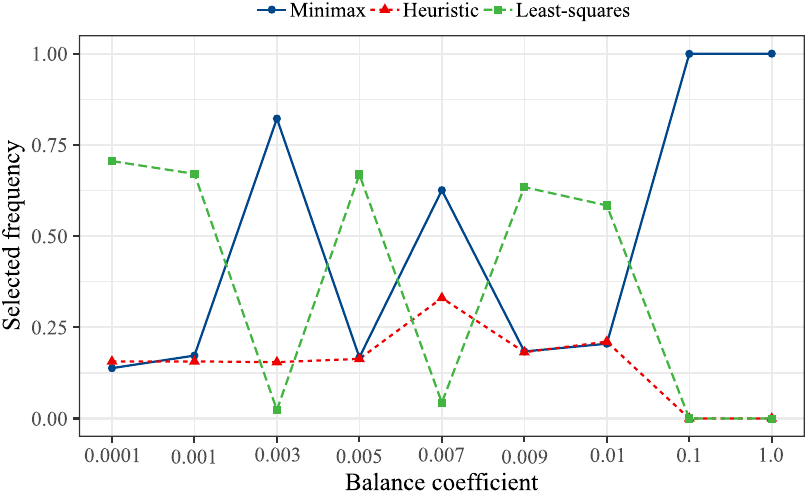}
		\label{fig:SelectedRatew.r.t.Gamma20k}}
	\hfil
	\subfloat[100K iterations]{
		\includegraphics[width=0.45\textwidth]{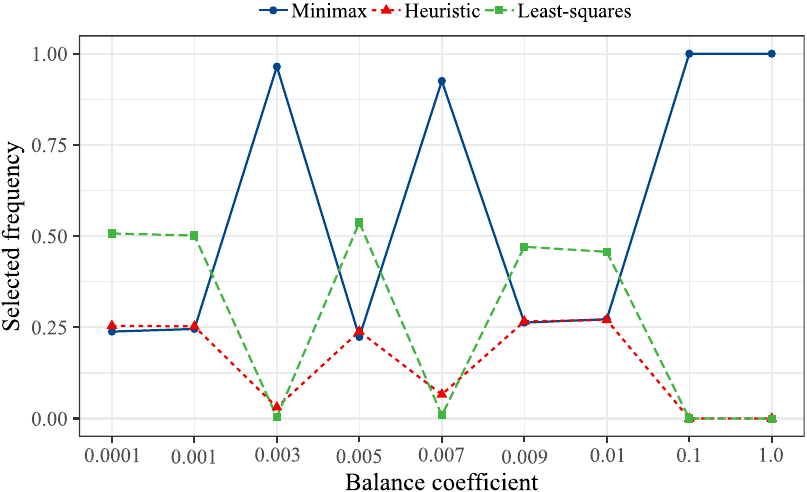}
		\label{fig:SelectedRatew.r.t.Gamma100k}} 
	\caption{{~{The} selected {frequency} of each operator in E-GANs with various balance coefficient{s}. (a) After training for 20K iterations. (b) After training for 100K iterations. {Here, the ``selected frequency" indicates the frequency with which individuals generated by a mutation operator are selected by the environment.}}}
	\label{fig:SelectedRatew.r.t.Gamma}
	\vspace{-0.5cm}
\end{figure*}

{We explore the effect of the diversity fitness function on the training process on the CIFAR-10 dataset. Fig.~\ref{fig:SelectedRatew.r.t.Gamma} shows that $F^{E-GAN}_d$ is such sensitive to $\gamma$ that a small change in value can lead to a reversal in the selected frequency of each operator. This sensitivity forces E-GAN to choose the $\gamma$ value carefully. Moreover, E-GAN's training process is accompanied by the risk of the vanishing gradients problem.} When the minimax objective dominates in the early stage (see $\gamma\in\{0.003, 0.007, 0.1, 1.0\}$ in Fig.~\subref*{fig:SelectedRatew.r.t.Gamma20k}), vanishing gradients is difficult to avoid (as shown in Fig.~\subref*{fig:SelectedRatew.r.t.Gamma100k}).
Due to the weakness of {the} minimax objective {in the early stage}, the offspring {generated by this objective} can be easily distinguished by the discriminator. However, $F^{E-GAN}_d$ gives unreasonably high scores to such distinguishable generators, which leads to the occurrence of {the} vanishing gradients {problem}. {In contrast, as shown in Table~\ref{tab:gamma1}, the FID values of IE-GANs varied regularly with the $\gamma_1$ values, indicating that $F^{IE-GAN}_d$ is not overly sensitive to $\gamma_1$. IE-GANs with various the $\gamma_1$ values all have meaningful results, indicating that $F^{IE-GAN}_d$ does not enhance the possibility of vanishing gradients.}

The effect of the diversity fitness function {on the generative performance} is demonstrated by Table~\ref{tab:FIDofFd}. {In this experiment, the} IE-GANs discard the crossover operator. {``w/o $F_d$" indicates that only $F_q$ is used for the evaluation.} {In terms of the generative performance, IE-GAN without the crossover operator {and without} $F^{IE-GAN}_d$ is equivalent to E-GAN without $F^{E-GAN}_d$.}
{The e}xperimental results show that $F^{E-GAN}_d$ limits and even reduces {the} performance. By simply removing this costly component, {the generative performance of E-GAN can be significantly improved}. $F^{IE-GAN}_d$ does not have such an obvious side effect. The degradation of IE-GAN is much {smaller} than {the degradation of} E-GAN.
The gain{s} of E-GAN due to {multiple populations} compensate for the negative effect of $F^{E-GAN}_d$. Moreover, even maintaining a population of eight individuals, E-GAN is worse than the method that simply uses {the} quality fitness function. IE-GAN using $F^{IE-GAN}$ can outperform IE-GAN using $F^{IE-GAN}_q$ because of the increase in {the} population.

\begin{table}[!t]
	\caption{Comparison of the Results of Various Diversity Fitness Functions (w/o Crossover)}
	\label{tab:FIDofFd}
	\centering
	\begin{tabular}{lcc}
		\toprule
		\makecell[l]{\textbf{Method}} & {\textbf{Final FID}} & {\textbf{Best FID}} \\
		\midrule
		{{IE-GAN / E-GAN ($\mu = 1$, w/o $F_d$)}} & {$\textbf{27.06}$} & {$27.06$} \\
		{IE-GAN ($\mu = 1$)} & {$27.90$} & {$27.90$} \\ 
		{IE-GAN ($\mu = 4$)} & {$27.15$} & {$\textbf{26.74}$} \\ 				
		{E-GAN ($\mu = 1$)} & {$29.38$} & {$29.38$} \\ 				
		{E-GAN ($\mu = 8$)} & {$28.45$} & {$-$} \\
		\bottomrule
	\end{tabular}
\end{table}

\begin{figure}[!t]
	\centering
	\includegraphics[width=0.485\textwidth]{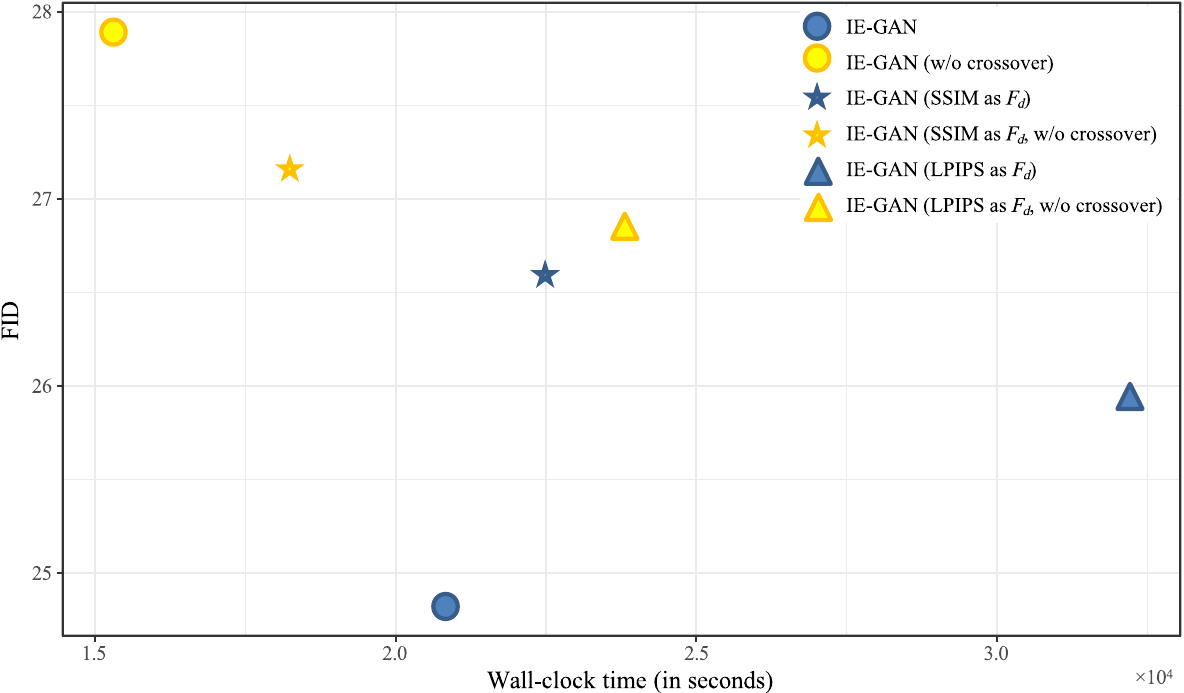}
	\caption{{{~The FID and wall-clock time of IE-GAN with different $F_d$ functions.}}}
	\label{fig:WCandFIDofFd}
	\vspace{-0.5cm}
\end{figure}

{In addition to validating $F^{IE-GAN}_d$, we have explored other possible options for $F_d$. Regarding the sample distance measure, in addition to the MAE, we also consider the structural similarity (SSIM) \cite{DBLP:journals/tip/WangBSS04} and learned perceptual image patch similarity (LPIPS) \cite{DBLP:conf/cvpr/ZhangIESW18}. Both are common metrics for measuring the similarity of images. The SSIM imitate human perception by focusing on the similarity of edges and textures. LPIPS extracts features from different layers of a pre-trained model to calculate the similarity of two image inputs. The literature indicates that LPIPS is more consistent with human perception than traditional methods \cite{DBLP:conf/cvpr/ZhangIESW18}. We use various image similarity measures as diversity fitness functions in the IE-GAN framework. The experiments compare the generative performance and runtime of IE-GAN with different alternative $F_d$ functions. The methods are trained for 100K iterations on the CIFAR-10 dataset. The results are reported in Fig.~\ref{fig:WCandFIDofFd}. For fairness, all IE-GANs are grid{-}searched for $\gamma_1 \in \{0, 0.001, 0.01, 0.05, 0.1, 0.5, 1\}$ and $\gamma_2 \in \{0, 0.0001, 0.001, 0.01, 0.1, 1\}$. The adopted hyperparameters are listed in Table~\ref{tab:gamm12ofFd}.}

\begin{table}[!t]
	\caption{{$\gamma_1$ and $\gamma_2$ Values Used by IE-GANs with Different $F_d$ Functions}}
	\label{tab:gamm12ofFd}
	\centering
	\begin{tabular}{lcc}
		\toprule
		\makecell[l]{\textbf{Method}} & {\textbf{$\gamma_1$}} & {\textbf{$\gamma_2$}} \\
		\midrule
		{\makecell[l]{IE-GAN \\ IE-GAN (w/o crossover)}} & {$ 0.05 $} & {$ 0.001 $} \\
		{\makecell[l]{IE-GAN (SSIM as $F_d$) \\ IE-GAN (SSIM as $F_d$, w/o crossover)}} & {$ 0.5 $} & {$ 0.001 $} \\
		{\makecell[l]{IE-GAN (LPIPS as $F_d$) \\ IE-GAN (LPIPS as $F_d$, w/o crossover)}} & {$ 0.05 $} & {$ 0.0001 $} \\
		\bottomrule
	\end{tabular}
	\vspace{-0.5cm}
\end{table}

{Fig.~\ref{fig:WCandFIDofFd} shows that, in general, a more advanced similarity metric improves the model's generative performance at the expense of time. In the absence of crossover, the IE-GAN {that uses the} LPIPS as $F_d$ has the lowest FID. Correspondingly, it has the longest runtime. With the addition of crossover, the LPIPS maintains an advantage over the SSIM in terms of the FID. However, the MAE has a better performance not only in terms of the runtime but also in terms of the FID. Compared to the LPIPS, the MAE has a significant time advantage; compared to the SSIM, the MAE also has better a generative performance. In summary, we finally choose the simple but useful MAE as $F^{IE-GAN}_d$.}

\vspace{-0.2cm}
\subsection{Crossover Operator Analysis}
In the implementation of PDERL\footnote{https://github.com/crisbodnar/pderl}, the $Q$-filtered behavior distillation crossover {method uses} a less capable network as the initial {offspring}, whereas we choose {a} better {network; this occurs due to} the differences between GANs and RL. The networks of GANs tend to be deeper, the generated samples are more time-sensitive, and the sample batch size is smaller to improve the efficiency.

\begin{figure}[!t]
	\centering
	\includegraphics[width=0.45\textwidth]{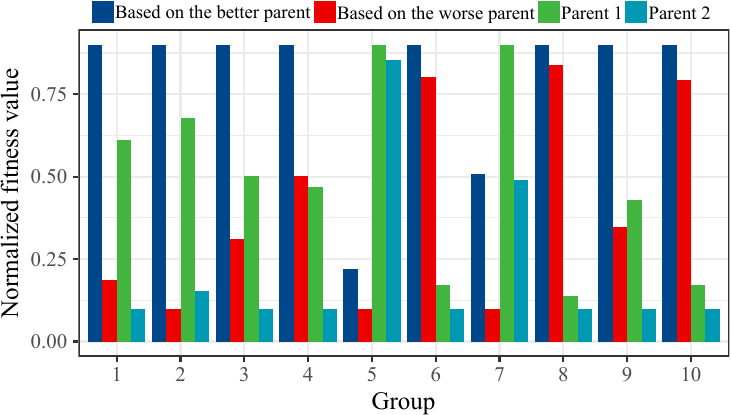}
	\caption{{~Normalized crossover performance.}}
	\label{fig:FIDofCross}
	\vspace{-0.25cm}
\end{figure}

To demonstrate {this}, we experimentally compare the two crossover operators, i.e., {the operator} based on {the} better parent and {the operator} based on {the} worse parent. The fitness of {the} offspring is a good indicator {that can be used} to measure {the} quality of {the} crossover operator. Fig.~\ref{fig:FIDofCross} plots ten randomly selected parent pairs{;} each set of bars includes the fitness {values} of {the} two parents{,} along with {the fitness values of} two types of crossover offspring. These values were normalized to [0.1, 0.9]. Not once did the crossover individual based on {the} worse parent {have a} higher fitness {value} than that based on {the} better parent, and it usually performs worse than {the} better parent {and} even {the} worse parent. 
It is difficult to assume that the contribution of the poorer individual {would} exceed that of the better {individual}.
The results in Table~\ref{tab:FIDofCross} are consistent with {our} expectations. {A b}etter operator can help the model to be better trained.

\begin{table}[!t]
	\caption{Comparison of the Results of Different Crossover Operators in Terms of the FID}
	\label{tab:FIDofCross}
	\centering
	\begin{tabular}{lcc}
		\toprule
		\makecell[l]{\textbf{Method}} & {\textbf{Final FID}} & {\textbf{Best FID}} \\
		\midrule
		{IE-GAN (based on the worse parent)} & {$25.92$} & {$25.92$} \\ 
		{IE-GAN (based on the better parent)} & {$\textbf{24.82}$} & {$\textbf{24.82}$} \\
		\bottomrule
	\end{tabular}
	\vspace{-0.5cm}
\end{table}

\vspace{-0.2cm}
\subsection{Validation on More Complex Datasets}
To demonstrate that our work can be applied to real-world image datasets, in addition to {the} CIFAR-10 dataset, we also conduct experiments on {the} CelebA dataset with {a} DCGAN architecture. In the face of this complex dataset, the network architecture is {somewhat} weak even with the enhanced number of channels. To maintain the equilibrium between $G$ and $D$, in this experiment{,} we restrict the discriminator update by setting $n_D=1$.
Six methods are tested: GAN, NS-GAN, LSGAN, GAN-Least-squares, E-GAN and our IE-GAN. Because the discriminator objective functions adopted by IE-GAN are originally used in GAN and NS-GAN, they are just GAN-Minimax and GAN-Heuristic for IE-GAN with $n_D=1$.
The quantitative results are shown in Table~\ref{tab:FIDofMethodsCelebA}. All methods {are unstable} on this dataset, so we only compare the best results. The FID value of the real samples is $48.96$.

\begin{table}[!t]
	\caption{Comparison of the Results of Different Methods on CelebA in Terms of the FID}
	\label{tab:FIDofMethodsCelebA}
	\centering
	\begin{tabular}{lc}
		\toprule
		\makecell[l]{\textbf{Method}} & {\textbf{Best FID}} \\
		\midrule
		{GAN-GP (GAN-Minimax) {\cite{DBLP:conf/nips/GoodfellowPMXWOCB14}}} & {$56.88$} \\
		{NS-GAN-GP (GAN-Heuristic) {\cite{DBLP:conf/iclr/ArjovskyB17}}} & {$54.58$} \\
		{LSGAN-GP \cite{DBLP:conf/iccv/MaoLXLWS17}} & {$376.79$} \\
		{GAN-Least-squares} & {$58.01$} \\
		{E-GAN {\cite{DBLP:journals/tec/WangXYT19}}} & {$59.66$} \\
		{IE-GAN (Ours)} & {$\textbf{52.582}$} \\
		\bottomrule
	\end{tabular}
	\vspace{-0.5cm}
\end{table}

It is clear from the observation{s} that LSGAN essentially fails to be trained and cannot generate meaningful samples. {Meanwhile}, GAN-Least-squares can be trained, but unlike on CIFAR-10, its performance is the worst {out of} all {the} single{-}mutation{-}operator methods. E-GAN, as always, does not benefit from multiple operators and performs the worst {out of all the} trainable methods. {O}ur IE-GAN achieves the best results, and its score is quite close to that of the true distribution.
{The experiments reveal that IE-GAN has good robustness.} It can achieve promising results on different datasets, {with} different network architectures, and {with} different operator performances. 

\vspace{-0.2cm}
\section{Conclusion}
\label{Conclusion}
In this paper, we verify that the evolutionary strategy of E-GAN, especially the diversity fitness function, is unreasonable. We propose a crossover operator that can be widely applied to evolutionary {GANs} and a more sensible fitness function. We unify them into a universal framework called IE-GAN and implement {this framework} based on E-GAN {and CDE-GAN}. Experiments demonstrate that our approach {provides a} better generative performance {and requires} less time {than previous methods}.
\vspace{-0.2cm}

\ifCLASSOPTIONcaptionsoff
\newpage
\fi

\bibliographystyle{IEEEtran}
\bibliography{Reference}

\end{document}